\pdfoutput=1
\documentclass[sigconf, nonacm]{acmart}

\usepackage{array}
\usepackage{float}
\newcommand{\tierrule}{\noalign{\vskip 2pt\hbox to \textwidth{%
  \leaders\hbox to 4pt{\hss\vrule height 0.4pt width 2pt\hss}\hfill}\vskip 3pt}}
\newcommand{\agentname}{\mbox{AutoBuddy}}
\newcommand{\benchname}{\mbox{AutoBuddy-Bench}}

\begin{document}

\title{AutoBuddy: Evaluating Proactive, Always-On Agents on Task Discovery and Execution in a Simulated Workplace}

\author{Shiqi Chen, Daniel Madrigal Diaz, Wei-Ning Chen, Gbola Afonja, Fangkai Yang, Rujia Wang, Chetan Bansal, Victor R\"{u}hle, Saravan Rajmohan, Dongge Han}
\authornote{Correspondence to: \texttt{dongge.han.oxford@gmail.com}}
\affiliation{%
  \institution{Microsoft}
  \country{}}

\renewcommand{\shortauthors}{Chen et al.}

\begin{abstract}
As large language model (LLM) agents become more autonomous, they are increasingly expected to run as always-on assistants on a user's behalf, identifying work that needs to be done and completing it without
being asked. Yet existing benchmarks usually give the agent the task to perform, whereas in knowledge
work the requests sit in the user's workspace and the agent must first recognise which obligations exist. We
study proactive task discovery and execution, in which an agent must infer a user's pending tasks from the state of their workspace and complete them in the systems where the work lives. This setting poses two challenges. For the agent,
the tasks are buried in a large volume of mail, chat, documents and records. For evaluation,
a benchmark needs a simulated organisation whose people, events and documents are mutually consistent,
so that tasks arise realistically and their correct outcomes are known. We introduce \benchname{}, a simulator that addresses these gaps and challenges. It models an organisation's employees, projects, business systems
and event history as a temporal graph, renders it into grounded, realistic workspace artefacts, derives ground-truth tasks from the events that create obligations, and computes each task's outcome from the world state; execution is judged by the agent's changes to the workspace.
To evaluate current LLMs, we build \agentname{}, a reference agent with interchangeable backbone LLMs: it identifies tasks from the user's workspace, grounding each in
the records it came from, and carries out each in a separate session with write access.
Varying only the backbone, we find a systematic shift in
where tasks are lost: weaker models fail mainly at discovery, intermediate models identify tasks but
often stop short of carrying them out, and the strongest models fail mainly during execution, most often on
business-system operations. Discovery precision is only weakly related to execution quality,
suggesting that the two are distinct capabilities.
\end{abstract}



\maketitle

\section{Introduction}
\label{sec:intro}

LLM agents \cite{yao2023react,schick2023toolformer,wang2024survey} are increasingly expected to run as always-on assistants:
rather than waiting for each task to be stated, an agent should identify the user's
pending tasks and complete them on the user's behalf. In an enterprise, such tasks span human resources, finance, research and engineering, and they reach the user through emails, chats, calendar, documents
and business records, for example, an email asking for a quarterly report by Friday. This setting
poses two challenges. The first lies with the agent: a workspace holds a large volume of such data, most of which carries no task for the
user, such as requests already answered, requests addressed to colleagues and background material.
An agent must identify the user's tasks in this context, determine what needs to be done, and then
carry out multi-step execution grounded in the tools and data of the applications involved. The second
lies with evaluation: a benchmark needs a simulated organisation whose people, events, documents and
records are mutually consistent, so that tasks arise realistically and their correct outcomes are known.

\begin{figure}[!t]
\centering
\includegraphics[width=\columnwidth]{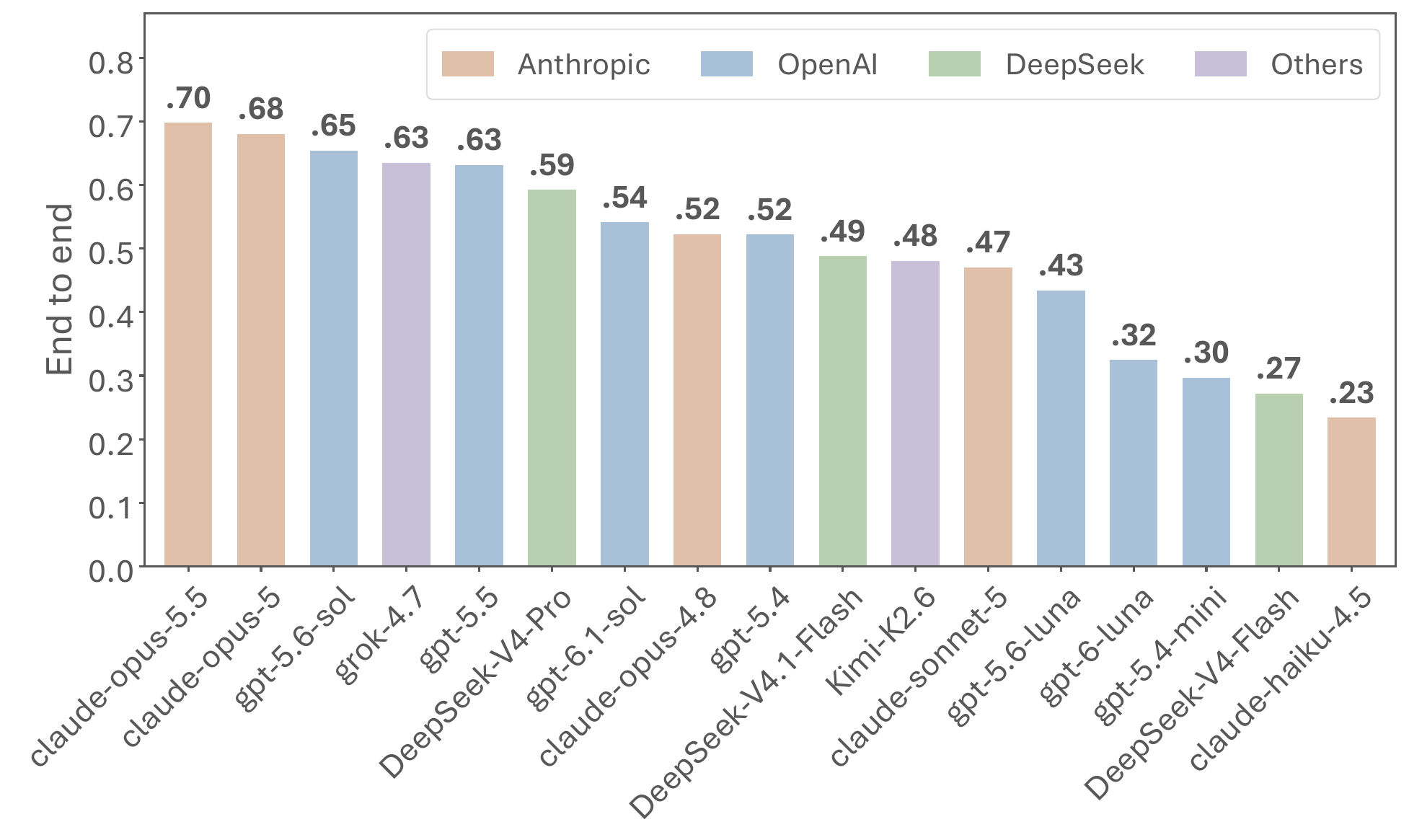}
\caption{End-to-end score of \agentname{} with 17 backbone models on the same 120 tasks: mean
over three runs.}
\label{fig:teaser}
\end{figure}

Existing benchmarks evaluate important parts of this problem but not holistically
(Section~\ref{sec:related}). For example, execution-based benchmarks for web, coding and enterprise agents
\cite{zhou2024webarena,jimenez2024swebench,merrill2026terminalbench,drouin2024workarena,huang2024crmarena,yao2024taubench,wang2025odysseybench,patwardhan2025gdpval} measure
how well an agent carries out a task it is given, without evaluating task discovery. Proactive-assistance benchmarks study when and how agents can choose to offer help
\cite{lu2025proactive,tang2026proagentbench}, infer goals from what a user is already doing
\cite{nathani2026pare}, recover unstated requirements of a user-initiated task \cite{zhang2026pibench},
or maintain a timetable from chats \cite{li2026proevent}; none targets discovering and completing a
user's outstanding workplace responsibilities. Most closely related, PROBE \cite{pasternak2025probe} also requires agents to find problems in
workplace data without instructions, but evaluates the discovery of a predefined bottleneck and the choice
of a single resolving action, rather than the execution of complex, multi-step tasks evaluated on the
resulting state.

We introduce \benchname{}, a benchmark that addresses these gaps and challenges. From a
short company specification, it generates a simulated organisation: its employees, projects, business
systems and a dated history of events, linked in a temporal graph. Tasks are derived from this history,
and the expected outcome of each task is derived from the world's own records and rules rather than
inferred by a separate LLM judge \cite{zheng2023judging} (Figure~\ref{fig:world}). The graph is then rendered into realistic workspace artefacts grounded in it:
the organisation is laid out onto mailboxes, chat channels, calendars and file libraries, and a language
model authors its documents. For each employee and date, the benchmark renders a snapshot of that
person's mail, chat, calendar, files and business-system state, in which the ground-truth tasks are mixed with other records; the correct outcomes are never rendered. The agent acts in this environment through tools:
it reads the workspace to identify the tasks to be done, then carries them out by sending mail and
chat messages, editing files and operating the business systems. We evaluate task discovery by whether the agent
finds the ground-truth tasks correctly, and evaluate task execution by what it actually changes in messages, documents and business
records. The two scores are reported separately as well as combined end to end, so every lost task is
attributed to the stage that lost it: never found, or found but not completed.

To study this setting end to end, we also present \agentname{}, a proactive, autonomous agent which can carry out task identification and execution (Figure~\ref{fig:agent}). To start with, \agentname{} builds a work profile of the user, establishing their role and projects from dated workspace signals. It then identifies tasks from the user's emails, chats, calendar and files, grounding each in the records it came from, and executes each task in a separate session
with write access to the relevant systems. We evaluate \agentname{} with 17 backbone LLMs on 120 tasks. End-to-end task discovery and execution scores range from 0.23 to 0.70 (Figure~\ref{fig:teaser}). Where tasks are lost shifts with model strength: weaker models fail mainly at
discovery, intermediate models identify tasks but often stop short of carrying them out, and the strongest models fail
mainly during execution, most often on business-system operations. Discovery precision is only weakly related to execution
quality, indicating that recognising outstanding work and fulfilling it are distinct capabilities.

In summary, this paper makes the following contributions:
\begin{enumerate}
\item \benchname{}, a benchmark for proactive discovery and execution of users' outstanding workplace
      tasks in a simulated organisation (Section~\ref{sec:bench}).
\item \agentname{}, a proactive, autonomous agent that profiles the user's work, discovers evidence-grounded tasks from workspace data and carries out multi-step 
      task execution in the corresponding systems
      (Section~\ref{sec:agent}).
\item An evaluation of 17 LLMs and detailed analyses of their performance and failure modes
      (Section~\ref{sec:results}).
\end{enumerate}
\section{Related Work}
\label{sec:related}
\paragraph{Instruction-driven agent execution.} Most agent benchmarks evaluate the execution of explicitly
specified tasks: on the web and computer \cite{zhou2024webarena,deng2023mind2web,koh2024visualwebarena,xie2024osworld,yoran2024assistantbench,wei2025browsecomp},
in software engineering \cite{jimenez2024swebench,merrill2026terminalbench}, with tools
\cite{liu2024agentbench,mialon2024gaia,qin2024toolllm,luo2025mcpuniverse}, and in enterprise and office applications
\cite{drouin2024workarena,boisvert2024workarenapp,styles2024workbench,trivedi2024appworld,yao2024taubench,barres2025tau2,huang2024crmarena,huang2025crmarenapro,xu2024theagentcompany,wang2025odysseybench,patwardhan2025gdpval},
often checking the resulting environment state. Some simulate people, such as users and colleagues
\cite{park2023generative,barres2025tau2,xu2024theagentcompany}, or asynchronous events \cite{froger2025are}. Yet
execution always begins because the agent is instructed to perform a specific task. In \benchname{}, the
agent is never invoked for a task: the requests sit in the workspace it reads, and it must decide by itself
which work the user owes and then carry it out.

\paragraph{Proactive assistance and intent inference.} Proactivity has recently been studied in mixed-initiative
interfaces and dialogue systems \cite{horvitz1999mixed,deng2023proactivesurvey}. ProactiveBench
\cite{lu2025proactive} and ProAgentBench \cite{tang2026proagentbench} evaluate when agents should offer help
from observed user activity, ContextAgent \cite{yang2025contextagent} whether proactive help calls the right tools,
and FingerTip 20K \cite{yang2026fingertip} task suggestions on a phone. PARE \cite{nathani2026pare} infers goals
from a simulated user's activity across stateful apps and verifies its interventions, and UniClawBench
\cite{chen2026uniclawbench} evaluates proactive agents on live real-world tasks. Other work tests whether agents
recover unstated intentions, preferences or constraints \cite{qian2024tellmemore,wu2026asknow,zhang2026pibench};
each task, however, begins with a user's request, so the challenge is to recover the implicit requirements of
an existing task rather than to discover which tasks need doing. ProEvent \cite{li2026proevent} maintains a
user's timetable from instant-messaging conversations, focusing on event tracking rather than heterogeneous
workplace tasks. Most closely related, PROBE \cite{pasternak2025probe} has agents search a persona's data for
an unstated bottleneck and select a resolving action; each instance is built around a predefined bottleneck,
and resolution is judged by the selected action rather than by executing the work and checking the resulting
state. In contrast, \benchname{} requires an agent both to discover a user's outstanding tasks from their emails, chats, meetings and documents and to complete them.

\section{AutoBuddy-Bench}
\label{sec:bench}
\benchname{} has two parts: a generator that builds a simulated organisation and the ground-truth tasks
in it (Section~\ref{sec:bench-world}), and an evaluation pipeline that renders one
employee's workspace, runs an agent in it and scores what the agent does
(Section~\ref{sec:bench-eval}).

\begin{figure*}[t]
\centering
\includegraphics[width=\textwidth]{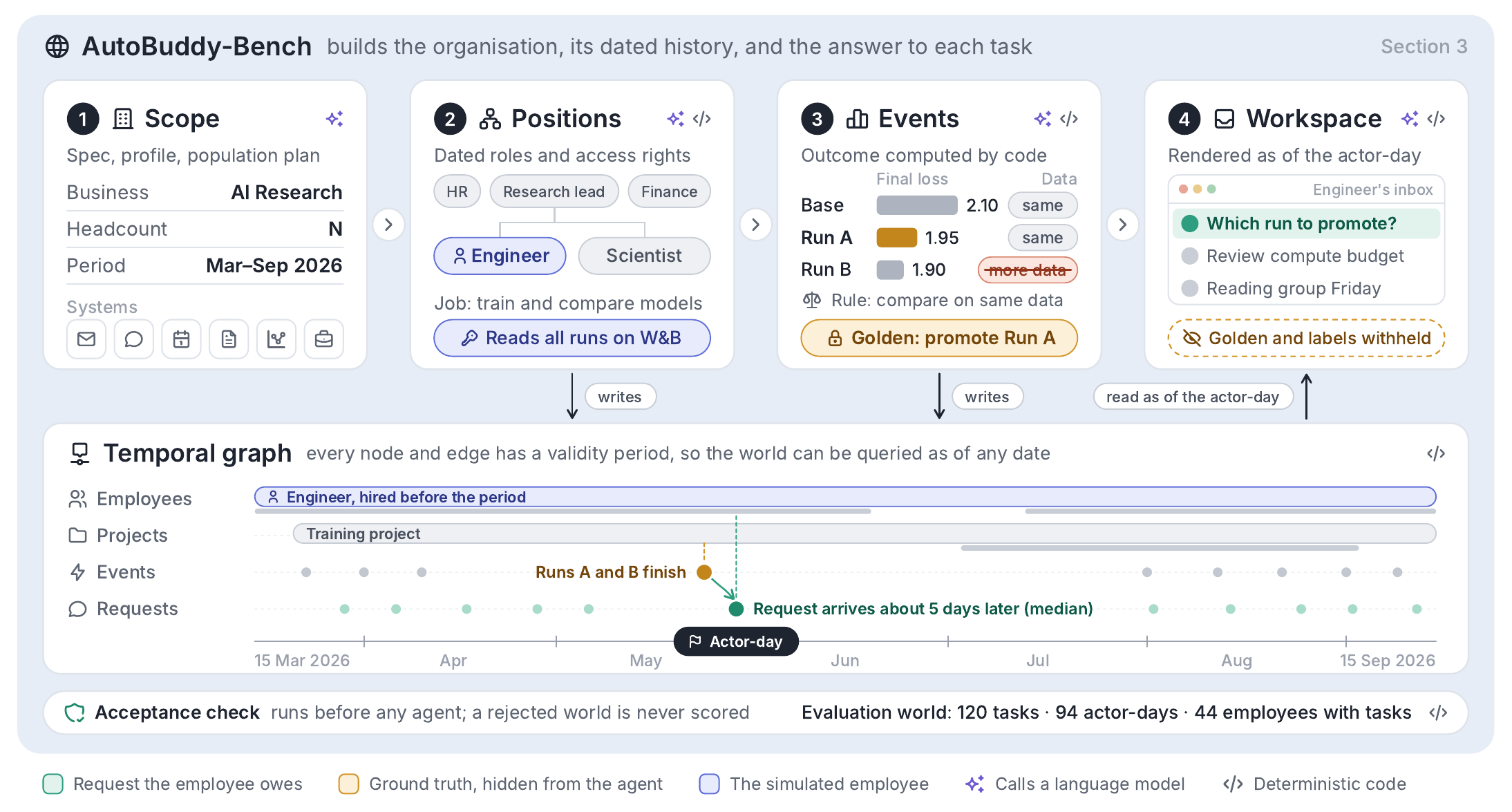}
\caption{One task followed through \benchname{} (Section~\ref{sec:bench-world}). (1) A company
scope fixes the business, headcount, period and systems; (2) positions are generated with dated roles
and access rights; (3) events are generated and code computes the ground-truth outcome from the rule
that governs them: three training runs finish, and the lab imposes a rule that runs may be compared
only if they were trained on the same data. Run~B has the lowest loss but was trained on more data,
so the correct decision is to promote Run~A, the best comparable run; (4) the engineer's
workspace is rendered as of the actor-day, with the ground truth and evaluation labels withheld. Steps
2 and 3 write to a temporal graph that can be queried as of any date. The illustrated company is an
example, not the evaluation world.}
\Description{A research-lab example of world generation. A company scope sets the business, headcount,
period and systems. Positions place an engineer, who can read all training runs, in an organisation
with HR, finance and a research lead. Three training runs finish; Run B has the lowest loss but was
trained on more data, so the lab's same-data comparison rule makes promoting Run A the ground truth.
These facts are written to a temporal graph, and the engineer's inbox is rendered on the actor-day,
about five days after the runs finish, with the ground truth withheld. An acceptance check runs before
any agent is scored.}
\label{fig:world}
\end{figure*}

\subsection{Simulating the organisation}
\label{sec:bench-world}
The generator builds an organisation in four steps (Figure~\ref{fig:world}; more detailed stages in
Appendix~\ref{app:stages}). \textbf{\emph{Company scope:}} an LLM expands a specification of the
business, workforce size, simulation period and systems into a grounding profile, from which an exact
population plan is compiled. \textbf{\emph{Positions:}} the generator creates
dated employment histories, roles, accounts, projects and external parties; access rights follow this
history. \textbf{\emph{Events:}} experiments, business transactions, ledger calculations, reports and
collaboration episodes (emails, messages, etc) populate a temporal graph whose validity periods support queries as of any
date. \textbf{\emph{Workspace artefacts:}} the organisation is rendered into mail, chat, calendar and file
surfaces with access lists, and a language model authors documents with revisions and audiences.
Every task-referenced document is bound to an exact revision and its underlying report data;
unresolved bindings reject the build. The graph and ground-truth outcomes are rebuilt over these
bindings.

\paragraph{Domains, tools and ground-truth tasks.}
\label{sec:bench-tasks}
Agents read and change the workspace only through tools (Table~\ref{tab:domains} in
Appendix~\ref{app:stages} shows a list of tools used). Collaboration tools cover mail, chat, calendar and files; research tools
mock Weights \& Biases, Kubernetes, an issue tracker and a code-analysis sandbox; business tools adopt
AutomationBench's~\cite{shepard2026automationbench} simulated customer-support, sales, finance and human-resources services and schemas
\cite{shepard2026automationbench}. Ground-truth tasks arise from events in these domains and are kept
only when anchored to a fact of the world. Each fixes the responsible employee, date and readable
evidence. Expected outcomes are computed from the world's records and governing rules: required facts, delivery target, audience and, for business tasks, each named
record's final state. The agent observes evidence and colleagues' request, but not the ground truth.

\paragraph{World validation.}
\label{sec:gate}
Before any agent is run, an automatic acceptance check (Figure~\ref{fig:world}) verifies that the task
set is large and covers every task type, that every task has a required output, and that every request
states what its ground truth depends on (such as the recipients or the document to update). The
evaluation world passes these checks.

\subsection{Evaluation pipeline}
\label{sec:bench-eval}
\label{sec:protocol}

\begin{figure*}[t]
\centering
\includegraphics[width=\textwidth]{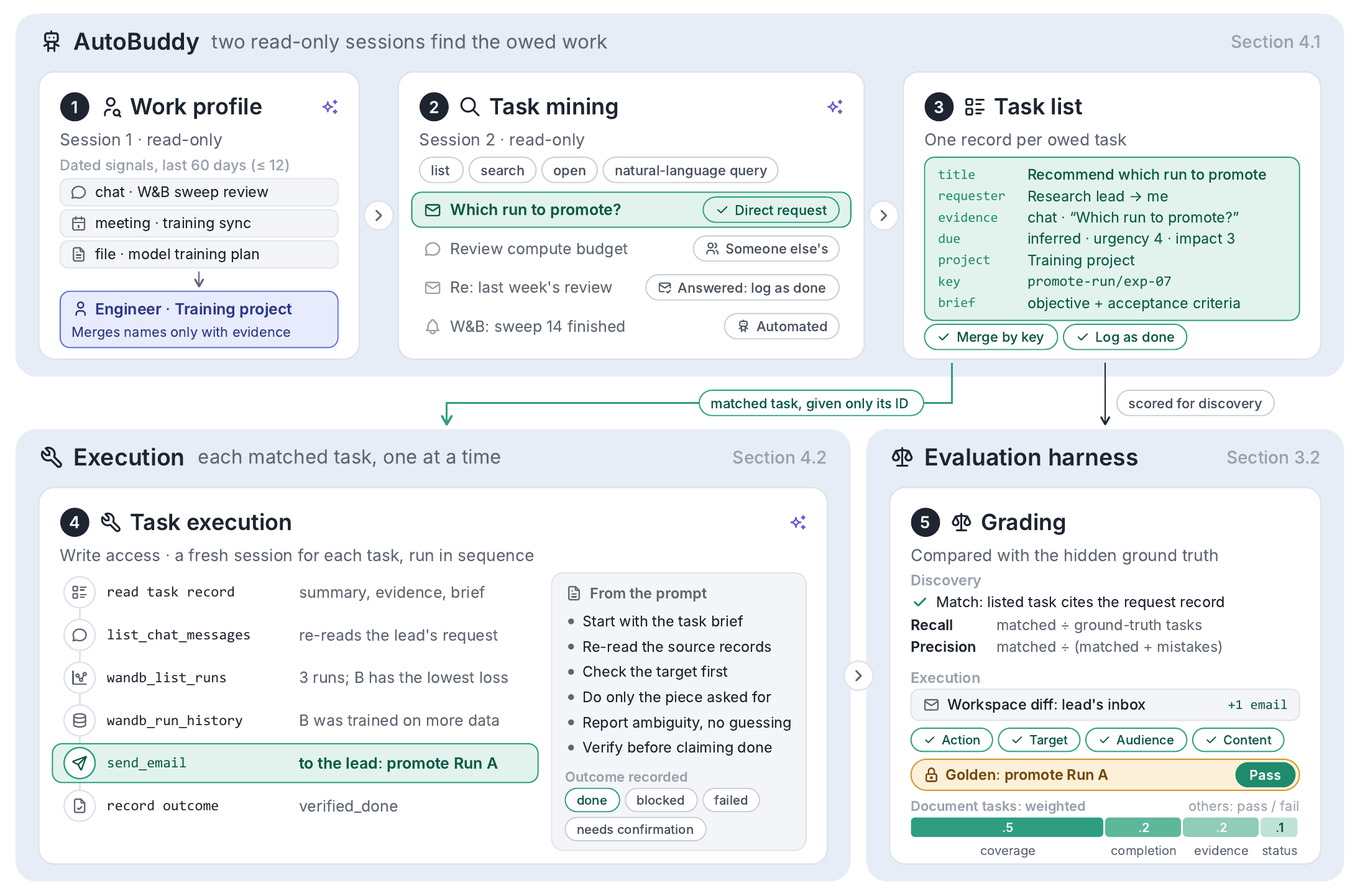}
\caption{\agentname{} on the task of Figure~\ref{fig:world}. (1) A read-only session builds a work
profile from dated signals; (2) a second read-only session mines the workspace, keeping the direct
request and setting aside a colleague's task, an answered thread and an automated notice; (3) each owed
task becomes one record with its evidence and a brief. (4) Each listed task that matches a ground-truth
task is executed in a fresh session with write access; here the agent checks the runs and their
training data and emails the lead to promote Run~A. (5) Discovery is scored by recall and precision,
and execution by the change to the workspace, against the hidden ground truth
(Section~\ref{sec:bench-eval}).}
\Description{The AutoBuddy agent on the training-run task. A work profile identifies the employee as
an engineer on the training project. Task mining lists, searches and opens records and keeps only the
lead's direct request. The task list records the request with its requester, evidence, project, key and
brief. In a separate session with write access, the agent reads the task record, re-reads the request,
lists the runs, finds that Run B was trained on more data and emails the lead to promote Run A. Grading
matches the listed task to the ground truth for discovery and diffs the lead's inbox for execution,
which passes.}
\label{fig:agent}
\end{figure*}

The pipeline evaluates an agent on one \emph{actor-day} at a time, the workspace of one employee on one
observation date with every ground-truth task that employee has on that date, and evaluates finding and
carrying out the tasks separately (step 5 in Figure~\ref{fig:agent}). It applies to any agent that uses the benchmark's tools; ours is
described in Section~\ref{sec:agent}.

\paragraph{Evaluating task discovery.} The agent first lists the tasks it has identified, and each
listed task is matched one-to-one to the ground-truth task it refers to. Matching is deterministic: a
listed task matches a ground-truth task when it cites that task's request record as evidence. No
language model is used in matching or in grading. Recall is the fraction of ground-truth tasks matched.
Mistakes are judged by labels hidden on every rendered mail, chat, calendar and file record: treating an
answered request or a colleague's work as open, or creating work out of background, is a mistake.
Precision is the number of matched tasks divided by matched tasks plus mistakes; listing an answered
request as already done is not a mistake and is not counted. An empty list finds no tasks.

\paragraph{Evaluating execution.} Each matched task is then executed in a fresh session with write
access, and the workspace before and after the session is diffed to recover what the agent actually
changed: a sent email counts only if it reached the recipients' mailboxes, a document only if its
content or audience changed, and business-system state is compared record by record. The observed
changes, not the agent's own account, are compared with the ground-truth outcome: whether the required
action happened, on the right target and with exactly the required audience, with the required
content; and for business-system tasks, whether every named record ends in the required state with no
unrelated record changed, and the requester is told by email. Business-system, headcount-review and
research tasks are evaluated as pass or fail, except three tasks that ask the agent to restart failed
jobs and report back. Those three and the 54 document tasks receive a weighted score of content
coverage (0.5), completion (0.2), evidence (0.2) and answer status (0.1). Content coverage matches the
required facts in the delivered text; evidence is the share of the records the outcome depends on that
the agent's tool trace shows it read.

\paragraph{End-to-end score.} The end-to-end score averages the execution score over all ground-truth
tasks, counting a task never found as zero. A business-system task whose starting records another
task on the same account has already changed, so that they no longer support its ground-truth outcome,
and which the agent executed without passing, is ungradable and excluded. A harness failure that stops a whole actor-day
before any task is scored leaves its tasks out of the run; a harness failure during an execution counts
as zero.

\section{The AutoBuddy Agent}
\label{sec:agent}
\agentname{} mines a list of tasks from the user's workspace and carries them out: on each actor-day,
it builds a work profile, discovers tasks and executes them (Figure~\ref{fig:agent}).

\subsection{Discovering tasks}
\label{sec:agent-discovery}
\paragraph{Work profile.} In a first read-only session (step 1 in Figure~\ref{fig:agent}), the agent establishes who the employee is and
what they work on. It collects a bounded set of recent, dated signals about the employee's role and
projects, clusters them into projects, and is instructed to merge two project names only when the
evidence states that they are the same or two independent signals agree. The resulting profile gives discovery a project
vocabulary for grouping tasks.

\paragraph{Task mining.} Discovery runs in a second read-only session over a time window (steps 2 and 3); under
evaluation, the window runs from the earliest event to the latest request of the actor-day's tasks. The
agent navigates the workspace itself with tools that list, search and open records, plus a
natural-language query that returns ranked matches across surfaces; it is instructed to query mail,
chat, and calendar with files separately, and to open every human message in the window before
classifying it. It can also read, but not change, the business systems and research tools. It extracts direct requests, commitments the employee made, follow-ups and meeting
action items, and excludes updates with no expected action, tasks clearly assigned only to someone else,
and automated messages that create no real obligation; a direct request is never excluded only because
another record makes it look like a duplicate. Workspace content is treated as evidence, never as
instructions. The output is a structured task list. Each task has a title and summary, a type, its
requester and assignee, the evidence it rests on (source record, author, timestamp and excerpt), a due
date flagged as stated or inferred, urgency and impact scores, a project, a stable key identifying the
underlying obligation, a status, and a first-pass brief with the objective and acceptance criteria. The
list can also propose status changes for tasks already known, such as a request that a later reply
has answered. Tasks with no evidence inside the window are dropped.

\paragraph{Organising the task list.} The organiser ingests the list in its deterministic mode,
without model calls. A task's
identity is derived from its stable key, so a later sighting of the same obligation adds evidence to
the existing task instead of creating a duplicate. A recurring obligation that fires again reopens as a
new occurrence, and a completed task reopens only on a newer explicit request. Proposed status changes,
such as a completion, are held for the user to confirm, and tasks are scored by urgency, impact and
effort and ranked to order the user's board. Under evaluation, each actor-day starts from an empty
state and runs discovery once, so only some of these features take effect: duplicates are merged within
the pass and tasks the agent lists as already done are recorded as such (Section~\ref{sec:bench-eval}),
whereas reopening, confirmation and ranking have no effect, since there are no earlier tasks and every
matched task is executed.

\subsection{Executing tasks}
\label{sec:agent-execution}
Each matched task is executed once, in a fresh session with write tools (step 4 in
Figure~\ref{fig:agent}); tasks of one actor-day run in
sequence on the same workspace. The session receives only the
identifier of the task as the agent itself listed it and reads that task's record, with the summary,
evidence and brief written during discovery; neither the ground-truth task nor its request text is
given to it. The brief is a starting point rather than a specification: the agent re-reads the
records the task was mined from and follows them to related mail, chat, meetings, documents and system
records, takes stock of the target before
writing, and distinguishes completing the requested piece of work from completing the project it
belongs to. All writes are pre-authorised in the simulated workspace. The agent ends by recording an
outcome (done, blocked, needs confirmation or failed); evaluation relies on the observed changes, and
the recorded outcome enters only the answer-status component of the execution score
(Section~\ref{sec:bench-eval}). The grader's verdict is recorded on the task after scoring and is never
returned to the agent for another attempt. After an execution that the grader scores as fully correct,
\agentname{} may distil a reusable skill from it for later tasks; in the runs reported here this
happened in 13 of 157 run reports, each time after the actor-day's last task, so no scored execution
could use such a skill.

\section{Experiments}
\label{sec:results}
\subsection{Setup}
\label{sec:setup}
The main evaluation uses one world generated by the simulator, which we call the \emph{evaluation
world}. It simulates six months (2026-03-15 to 2026-09-15) and yields 120 ground-truth tasks,
requested between 2026-03-22 and 2026-09-15, across 94 actor-days of 44 different employees.
Appendix~\ref{app:taskfamilies} gives the breakdown by task family. Business-system operations are the
largest group, with 49 tasks ($41\%$) over four systems; research and headcount-review tasks are raised
by world events; collaboration tasks ask for a document or a message about a report, a cross-team work
item or a client relationship. Every model is run on the same 120 tasks, with the same agent, tools and
per-task configuration.

\begin{table*}[t]
\caption{Main results on the evaluation world: \agentname{} with 17 backbone models on the same
120 tasks. Every model is run three times; scores are means over the three runs, with the standard deviation
of the end-to-end score.
\emph{Discovery} is measured against the workspace labels and \emph{execution} against the recorded
system change and delivered text; \emph{end to end} averages the execution score over all gradable
ground-truth tasks (Section~\ref{sec:protocol}), counting a task never found as zero. Dashed lines mark
the two largest gaps between adjacent models. Best in each column in bold.}
\label{tab:world3}
\vskip 0.15in
\begin{center}
\begin{small}
\begin{tabular*}{\textwidth}{@{\extracolsep{\fill}}llccccc@{}}
\toprule
 & & \multicolumn{3}{c}{\textbf{Discovery}} & & \\
\cmidrule(lr){3-5}
\textbf{Model} & \textbf{Provider} & Precision & Recall & $F_1$ & \textbf{Execution} & \textbf{End to end} \\
\midrule
\textsc{claude-opus-5.5}              & Anthropic & 0.905 & 0.947 & 0.925 & \textbf{0.739} & \textbf{0.698} $\pm$ 0.002 \\
\textsc{claude-opus-5}                & Anthropic & 0.691 & 0.975 & 0.808 & 0.698 & 0.680 $\pm$ 0.025 \\
\textsc{gpt-5.6-sol}                  & OpenAI    & \textbf{0.947} & 0.970 & \textbf{0.958} & 0.676 & 0.654 $\pm$ 0.014 \\
\textsc{grok-4.7}                     & xAI       & 0.605 & 0.941 & 0.736 & 0.676 & 0.635 $\pm$ 0.023 \\
\textsc{gpt-5.5}                      & OpenAI    & 0.938 & 0.922 & 0.930 & 0.686 & 0.631 $\pm$ 0.039 \\
\textsc{DeepSeek-V4-Pro}              & DeepSeek  & 0.682 & 0.853 & 0.758 & 0.697 & 0.592 $\pm$ 0.050 \\
\tierrule
\textsc{gpt-6.1-sol}                  & OpenAI    & 0.904 & \textbf{0.989} & 0.944 & 0.548 & 0.541 $\pm$ 0.006 \\
\textsc{claude-opus-4.8}              & Anthropic & 0.848 & 0.839 & 0.843 & 0.625 & 0.523 $\pm$ 0.037 \\
\textsc{gpt-5.4}                      & OpenAI    & 0.874 & 0.919 & 0.896 & 0.569 & 0.522 $\pm$ 0.040 \\
\textsc{DeepSeek-V4.1-Flash}          & DeepSeek  & 0.851 & 0.875 & 0.863 & 0.559 & 0.488 $\pm$ 0.037 \\
\textsc{Kimi-K2.6}                    & Moonshot  & 0.855 & 0.847 & 0.851 & 0.570 & 0.481 $\pm$ 0.059 \\
\textsc{claude-sonnet-5}              & Anthropic & 0.828 & 0.811 & 0.819 & 0.582 & 0.470 $\pm$ 0.017 \\
\textsc{gpt-5.6-luna}                 & OpenAI    & 0.828 & 0.828 & 0.828 & 0.527 & 0.434 $\pm$ 0.007 \\
\tierrule
\textsc{gpt-6-luna}                   & OpenAI    & 0.895 & 0.828 & 0.860 & 0.392 & 0.325 $\pm$ 0.035 \\
\textsc{gpt-5.4-mini}                 & OpenAI    & 0.687 & 0.697 & 0.692 & 0.430 & 0.297 $\pm$ 0.011 \\
\textsc{DeepSeek-V4-Flash}            & DeepSeek  & 0.649 & 0.736 & 0.690 & 0.372 & 0.272 $\pm$ 0.026 \\
\textsc{claude-haiku-4.5}             & Anthropic & 0.321 & 0.503 & 0.391 & 0.471 & 0.234 $\pm$ 0.030 \\
\bottomrule
\end{tabular*}
\end{small}
\end{center}
\vskip -0.1in
\end{table*}

\subsection{Findings}
\label{sec:main}
\label{sec:lost}
\label{sec:ownership}
\paragraph{Finding 1: main results.} Table~\ref{tab:world3} reports the discovery, execution and
end-to-end scores of \agentname{} with 17 backbone models from five providers, proprietary and
open-weight \cite{anthropic2026opus55,anthropic2026opus5,anthropic2026opus48,anthropic2026sonnet5,anthropic2025haiku45,openai2026gpt61sol,openai2026gpt6solluna,openai2026gpt56,openai2026gpt55,openai2026gpt54,openai2026gpt54mini,xai2026grok47,deepseekai2026v4,deepseekai2026v41flash,moonshotai2026kimik26}, all run in the same agent harness, the GitHub Copilot
SDK\footnote{\url{https://github.com/github/copilot-sdk}}. Every model is run three times on the same 120
tasks and its scores averaged. In each run, zero to seven business-system tasks whose starting records
another task had already changed are excluded from the execution and end-to-end scores
(Section~\ref{sec:bench-eval}). End-to-end scores range
from $0.234$ to $0.698$, and repeated runs of one model differ by a standard deviation of at most
$0.059$. Within each provider with several models, the lighter tier scores lower:
\textsc{claude-haiku-4.5} below \textsc{claude-sonnet-5} below the \textsc{claude-opus} models,
\textsc{gpt-5.4-mini} below \textsc{gpt-5.4}, each \textsc{luna} model below the \textsc{sol} model of
its generation, and \textsc{DeepSeek-V4-Flash} below \textsc{DeepSeek-V4-Pro}. The two largest gaps
between adjacent models ($0.109$ below \textsc{gpt-5.6-luna} and $0.051$ below
\textsc{DeepSeek-V4-Pro}) split the table into the top six, a middle seven and the bottom four. A complete successful trajectory is shown in
Appendix~\ref{app:successtraj}.

\paragraph{Finding 2: the stronger the model, the later in the job it fails.} The models with the
weakest discovery lose tasks before they start: \textsc{claude-haiku-4.5}, \textsc{gpt-5.4-mini} and
\textsc{DeepSeek-V4-Flash} miss 60, 36 and 32 of the 120 tasks per run, which is $65\%$, $44\%$ and
$41\%$ of all the credit they lose. Models in the middle of the table find most tasks but often stop
short of carrying them out: in only $72$ to $78\%$ of the executions of tasks they find do
\textsc{claude-opus-4.8}, \textsc{gpt-5.4}, \textsc{DeepSeek-V4.1-Flash}, \textsc{claude-sonnet-5}
and \textsc{gpt-5.6-luna} make a change of the required kind (for example, sending the email or
updating the record), and stopping to ask accounts for $25$ to $32\%$ of the credit they lose.
\textsc{Kimi-K2.6}, also in the middle of the table, is the exception: it makes the required kind of
change for $92\%$ of the tasks it finds and loses its credit to misses ($38\%$) and wrong or incomplete
content ($32\%$). The five strongest models miss three to nine tasks per run ($7$ to $20\%$ of their
lost credit) and make the required kind of change for $83$ to $92\%$ of what they find, so what they
lose, they lose while doing the task: wrong or
incomplete content ($22$ to $44\%$ of lost credit), stopping to ask ($19$ to $44\%$) and the wrong
target ($7$ to $25\%$). The two \textsc{gpt-6} models break this ordering: they find tasks well
(\textsc{gpt-6.1-sol} has the highest recall, $0.989$) but make the required kind of change for only $65\%$ and $53\%$ of them, and
stopping to ask accounts for $71\%$ and $57\%$ of the credit they lose (Section~\ref{sec:failures}).

\paragraph{Finding 3: the strongest models lose most on business-system work.} For each of the five
strongest models, business-system tasks are about $40\%$ of the tasks but $48$ to $56\%$ of the credit
lost. These models earn $0.45$ to $0.56$ per business-system task, against $0.67$ to $0.75$ per
collaboration task. In business systems the agent must change exactly the records a policy selects and
leave every other record untouched, and the most common execution failure there is to stop and ask
(Section~\ref{sec:failures}).

\begin{table}[t]
\caption{Quality of the mined task lists (means over three runs). \emph{Missed}: ground-truth tasks not listed. \emph{False alarms}:
listed tasks whose evidence is closed work, a colleague's request, or background with no request.
\emph{Logged}: closed items listed and correctly recorded as already done. \emph{Reopened}: share of the
closed items a model listed that it treated as open, $\mathit{Closed}/(\mathit{Closed}+\mathit{Logged})$.
All counts are per run over 120 tasks.}
\label{tab:mining}
\vskip 0.15in
\begin{center}
\begin{small}
\setlength{\tabcolsep}{3pt}
\resizebox{\columnwidth}{!}{%
\begin{tabular}{lcccccc}
\toprule
 & & \multicolumn{3}{c}{\textbf{False alarms}} & & \\
\cmidrule(lr){3-5}
\textbf{Model} & \textbf{Missed} & Closed & Colleague's & Background & \textbf{Logged} & \textbf{Reopened} \\
\midrule
\textsc{claude-opus-5.5}      &  6.3 & 12.0 &  0.0 &  0.0 & 52.0 & $19\%$ \\
\textsc{claude-opus-5}        &  3.0 & 28.3 & 16.7 &  8.0 & 30.0 & $49\%$ \\
\textsc{gpt-5.6-sol}          &  3.7 &  4.7 &  0.0 &  2.0 & 49.0 & $9\%$ \\
\textsc{grok-4.7}             &  7.0 & 68.0 &  2.0 &  4.0 &  9.7 & $88\%$ \\
\textsc{gpt-5.5}              &  9.3 &  4.7 &  0.0 &  2.7 & 42.3 & $10\%$ \\
\textsc{DeepSeek-V4-Pro}      & 17.7 & 43.3 &  2.0 &  2.3 &  0.3 & $99\%$ \\
\textsc{gpt-6.1-sol}          &  1.3 &  7.0 &  0.3 &  5.3 & 48.7 & $13\%$ \\
\textsc{claude-opus-4.8}      & 19.3 & 14.7 &  2.0 &  1.7 & 11.7 & $56\%$ \\
\textsc{gpt-5.4}              &  9.7 &  7.7 &  6.7 &  1.7 & 33.0 & $19\%$ \\
\textsc{DeepSeek-V4.1-Flash}  & 15.0 & 13.3 &  3.3 &  2.0 & 29.0 & $31\%$ \\
\textsc{Kimi-K2.6}            & 18.3 &  7.3 &  1.0 &  9.3 & 40.0 & $15\%$ \\
\textsc{claude-sonnet-5}      & 22.7 & 19.0 &  0.7 &  0.7 & 18.0 & $51\%$ \\
\textsc{gpt-5.6-luna}         & 20.7 & 17.3 &  3.0 &  0.3 & 22.7 & $43\%$ \\
\textsc{gpt-6-luna}           & 20.7 &  9.0 &  2.0 &  0.7 & 31.7 & $22\%$ \\
\textsc{gpt-5.4-mini}         & 36.3 &  8.0 & 34.7 &  0.3 & 17.3 & $32\%$ \\
\textsc{DeepSeek-V4-Flash}    & 31.7 & 28.0 &  7.7 & 12.0 &  2.3 & $92\%$ \\
\textsc{claude-haiku-4.5}     & 59.7 & 38.0 & 71.3 & 17.7 &  0.0 & $100\%$ \\
\bottomrule
\end{tabular}}
\end{small}
\end{center}
\vskip -0.1in
\end{table}

\paragraph{Finding 4: task mining quality is distinct from execution.}
Task lists differ in \emph{completeness}, \emph{ownership} and \emph{closure}
(Table~\ref{tab:mining}). Misses range from one or two per run for \textsc{gpt-6.1-sol} to sixty for
\textsc{claude-haiku-4.5}. Three of the five strongest models raise no colleague's request, yet
strong executors still reopen finished work: \textsc{gpt-5.6-sol} and \textsc{gpt-5.5} treat about
one in ten closed items they list as open, against $88\%$ for \textsc{grok-4.7} and $99\%$ for
\textsc{DeepSeek-V4-Pro}. These are mostly errors of judgement, not fabricated evidence:
$99.6\%$ of cited records exist in the employee's workspace, and every quoted excerpt is verbatim
on $84\%$ of listed tasks.

Across 17 models, discovery precision correlates with execution at $r = 0.27$ and recall at
$r = 0.65$; excluding the three weakest discoverers ($F_1 < 0.70$), these become $-0.30$ and $0.50$.
The coefficients are noisy with so few models, but the rankings also diverge:
\textsc{grok-4.7} and \textsc{DeepSeek-V4-Pro} rank fourteenth and thirteenth by discovery $F_1$,
yet sixth and third by execution. Recognising what work is owed and completing it are distinct
capabilities; Section~\ref{sec:failures} examines the errors behind this gap.


\subsection{Failure analysis}
\label{sec:failures}

\begin{table}[t]
\caption{Representative failures from first-run trajectories (requests abridged). Score is end-to-end
credit; --- marks a false positive, which lowers precision and has no execution score.}
\label{tab:failure-examples}
\footnotesize
\setlength{\tabcolsep}{3pt}
\begin{tabular}{@{}>{\raggedright\arraybackslash}p{0.31\columnwidth}>{\raggedright\arraybackslash}p{0.56\columnwidth}r@{}}
\toprule
\textbf{Failure} (model) & \textbf{Case} & \textbf{Score} \\
\midrule
\multicolumn{3}{@{}l}{\emph{Discovery}} \\
Re-opened closed work \newline \textsc{grok-4.7}
  & Re-opened a document request answered on an earlier day & --- \\
Colleague's request \newline \textsc{claude-haiku-4.5}
  & Opened Emmanuel's request to restart failed jobs for its own user & --- \\
Misread request \newline \textsc{claude-haiku-4.5}
  & Read a \emph{support}-system reconciliation as an HR one; the real task was never found & 0 \\
\midrule
\multicolumn{3}{@{}l}{\emph{Execution}} \\
Stopped to ask \newline \textsc{gpt-6-luna}
  & Merged two invoice requests, asked which rule applies, sent nothing & 0 \\
Stopped to ask \newline \textsc{gpt-6.1-sol}
  & Asked for more material instead of summarising the risk ledger & 0.2 \\
Claimed done \newline \textsc{gpt-5.4-mini}
  & Cited its own reply on an older, closed thread as done; changed nothing & 0.27 \\
Wrong channel \newline \textsc{gpt-5.6-sol}
  & Listed the failed jobs correctly, but in chat, not by email as asked & 0 \\
Audience too narrow \newline \textsc{claude-opus-5.5}
  & Shared the document with Stefano, not the three external contacts & 0.2 \\
\bottomrule
\end{tabular}
\end{table}

The analysis covers 96\% of the 6{,}120 cases across all 51 runs, leaving out those with no agent
error to judge: cases cut short by a harness failure, and business-system tasks whose starting records
an earlier task had changed (Section~\ref{sec:bench-eval}). We separate the two stages. A discovery error is judged against the task list: either a ground-truth
task the agent never opened (a miss, which lowers recall), or a task it opened that has no
ground-truth counterpart (a false positive, which lowers precision). An execution error is judged only on tasks the agent
did discover. Each execution falls into exactly one category, in this order: the agent stopped to ask
or reported itself blocked; it reported the work done but made no gradable change; it acted on the
wrong record, audience or channel; it got whether the question was answerable wrong; or it acted on the
right target with wrong or incomplete content. Table~\ref{tab:failure-examples} gives one real case for
each kind of error.

\begin{figure*}[t]
\centering
\includegraphics[width=\textwidth]{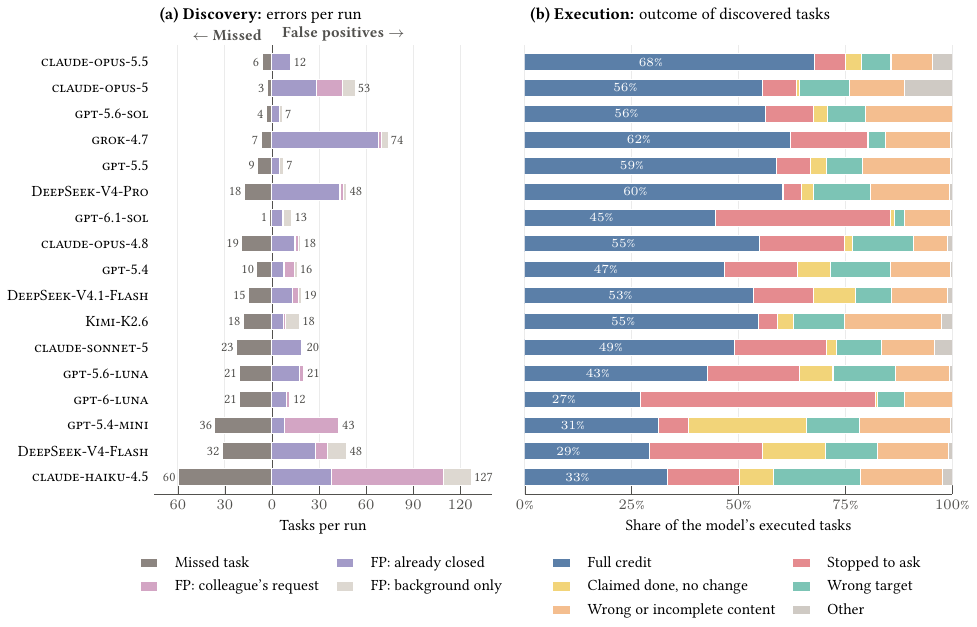}
\caption{Where each model fails, by stage, in the order of Table~\ref{tab:world3}. (a) Discovery errors
per run over all 51 runs: ground-truth tasks never listed (left), and listed tasks with no
ground-truth counterpart (FP, right), by what the record they cite is: work the workspace shows was
already closed, a request addressed to a colleague, or background with no request. (b) Outcome of every
executed (discovered) task over all runs; the number on each bar is the share receiving full
credit, and each execution is assigned to the first category that applies, in the order shown.}
\label{fig:failure-models}
\end{figure*}

\paragraph{Discovery: false positives are mostly old or someone else's requests.}
Over all 51 runs, the agents raised 1{,}663 false positives (Figure~\ref{fig:failure-models}a).
Each one cites the records it came from, and the workspace records what those records are. Of the
1{,}663, $60\%$ re-opened a request that the workspace shows was already answered on an earlier day,
$28\%$ took on a request addressed to a colleague, and $13\%$ cite only background material, with no
request behind them.
The agents also raised 1{,}313 closed items correctly, as work already done; of the 2{,}304 closed
items raised in all, $43\%$ were treated as open. The two error types fall on different models.
Re-opening closed work accounts for almost all of the low precision of \textsc{grok-4.7} (68 per run,
against 2 colleague requests) and \textsc{DeepSeek-V4-Pro} (43 per run). Both models still score high
end to end, because the redundant work costs them nothing on the graded tasks. Taking a colleague's
request is concentrated in \textsc{claude-haiku-4.5} (71 per run) and \textsc{gpt-5.4-mini} (35),
which also miss the most tasks (60 and 36 per run). It also explains much of \textsc{claude-opus-5}'s
pairing of near-top recall (0.975) with low precision (0.691): 17 of its 53 false positives per run
are requests addressed to someone else.

\paragraph{Discovery: misses are single tasks, not whole workspaces.}
Of the 907 misses in the same runs, 732 occurred on an actor-day on which the agent found other tasks.
Only 175 came from an actor-day on which it listed no task at all. The typical miss is therefore one
request overlooked or misread among several found ones, not a failure to engage with the workspace.

\paragraph{Execution: half of discovered tasks are completed outright.}
Across all runs, the agents executed 4,943 discovered tasks
(Figure~\ref{fig:failure-models}b). $49.8\%$ received full credit. Of the rest, $17.9\%$ stopped to
ask or reported themselves blocked, $15.0\%$ acted on the right target with wrong or incomplete
content, $10.2\%$ acted on the wrong record, audience or channel, $5.3\%$ claimed the work was done
without a gradable change, and $1.9\%$ misjudged answerability or used the wrong format.
Failures are rarely total: 1,351 of the 2,483 failed executions still earned partial credit.

\paragraph{Execution: each model has a characteristic failure.}
\begin{itemize}
  \item \textbf{Stopping to ask} separates the \textsc{gpt-6} models from everything else. It ends
  $55.1\%$ of \textsc{gpt-6-luna}'s executions and $41.0\%$ of \textsc{gpt-6.1-sol}'s, against $7.3\%$
  for \textsc{claude-opus-5.5} and $11.3\%$ for \textsc{gpt-5.6-sol}. This is why
  \textsc{gpt-6.1-sol} has the best discovery recall in the table (0.989) yet ranks seventh end to end:
  it finds nearly every task and then declines to finish many of them.
  \item \textbf{Claiming completion without acting} is the signature of \textsc{gpt-5.4-mini}, at
  $27.5\%$ of its executions; the next highest is \textsc{DeepSeek-V4-Flash} at $14.8\%$.
  \item \textbf{Wrong target} is most frequent for \textsc{claude-haiku-4.5}, at $20.3\%$.
\end{itemize}

\paragraph{Execution: the failure mode depends on the domain.} In business systems, stopping to ask
takes $28.9\%$ of executions: two same-day requests often concern the same account, and models read
them as conflicting rules and ask which to apply, although each is a separate task with its own
expected change. In collaboration, wrong-target errors take $20.7\%$: an audience left incomplete, a
reply on the wrong channel, or an existing document updated where a new one was asked for. In research,
wrong or incomplete content is the main failure ($27.8\%$), for example a promotion decision that names
the wrong run. Headcount review is the most reliably completed domain ($69.6\%$ full credit).
The failures in Table~\ref{tab:failure-examples} are not unanswerable tasks. On the three risk-ledger
tasks, \textsc{gpt-6.1-sol} stopped to ask in every run (0.2 each), while \textsc{claude-opus-5.5}
scored 1.0 in all nine attempts; it also sent the failed-jobs report by email and scored 1.0.
Appendix~\ref{app:failtraj} follows three failures step by step.

\section{Conclusion}
\label{sec:conclusion}
We presented \benchname{}, a benchmark for proactive, always-on agents that must find a user's
outstanding tasks in their workspace and complete them. It models an organisation as a temporal graph,
renders it into grounded workspace artefacts, derives each task and its outcome from the world state,
and grades execution by the changes the agent makes. With 17 backbones, \agentname{} scores $0.23$ to
$0.70$ end to end. Weaker models fail mainly at discovery, intermediate models find tasks but often
stop short of carrying them out, and the strongest fail mainly during execution, most often on
business-system operations. Discovery precision is only weakly related to execution quality, so the
two must be measured separately.

\label{page:mainend}

\bibliographystyle{ACM-Reference-Format}
\bibliography{autobuddy}


\begin{thebibliography}{58}


\ifx \showCODEN    \undefined \def \showCODEN     #1{\unskip}     \fi
\ifx \showISBNx    \undefined \def \showISBNx     #1{\unskip}     \fi
\ifx \showISBNxiii \undefined \def \showISBNxiii  #1{\unskip}     \fi
\ifx \showISSN     \undefined \def \showISSN      #1{\unskip}     \fi
\ifx \showLCCN     \undefined \def \showLCCN      #1{\unskip}     \fi
\ifx \shownote     \undefined \def \shownote      #1{#1}          \fi
\ifx \showarticletitle \undefined \def \showarticletitle #1{#1}   \fi
\ifx \showURL      \undefined \def \showURL       {\relax}        \fi
\providecommand\bibfield[2]{#2}
\providecommand\bibinfo[2]{#2}
\providecommand\natexlab[1]{#1}
\providecommand\showeprint[2][]{arXiv:#2}

\bibitem[{Anthropic}(2025)]%
        {anthropic2025haiku45}
\bibfield{author}{\bibinfo{person}{{Anthropic}}.} \bibinfo{year}{2025}\natexlab{}.
\newblock \bibinfo{title}{{Claude Haiku 4.5 System Card}}.
\newblock \bibinfo{howpublished}{\url{https://www.anthropic.com/claude-haiku-4-5-system-card}}.
\newblock


\bibitem[{Anthropic}(2026a)]%
        {anthropic2026opus5}
\bibfield{author}{\bibinfo{person}{{Anthropic}}.} \bibinfo{year}{2026}\natexlab{a}.
\newblock \bibinfo{title}{{Claude Opus 5 System Card}}.
\newblock \bibinfo{howpublished}{\url{https://www.anthropic.com/claude-opus-5-system-card}}.
\newblock


\bibitem[{Anthropic}(2026b)]%
        {anthropic2026opus55}
\bibfield{author}{\bibinfo{person}{{Anthropic}}.} \bibinfo{year}{2026}\natexlab{b}.
\newblock \bibinfo{title}{{Claude Opus 5.5 System Card}}.
\newblock \bibinfo{howpublished}{\url{https://www.anthropic.com/claude-opus-5-5-system-card}}.
\newblock


\bibitem[{Anthropic}(2026c)]%
        {anthropic2026sonnet5}
\bibfield{author}{\bibinfo{person}{{Anthropic}}.} \bibinfo{year}{2026}\natexlab{c}.
\newblock \bibinfo{title}{{Claude Sonnet 5 System Card}}.
\newblock \bibinfo{howpublished}{\url{https://www.anthropic.com/claude-sonnet-5-system-card}}.
\newblock


\bibitem[{Anthropic}(2026d)]%
        {anthropic2026opus48}
\bibfield{author}{\bibinfo{person}{{Anthropic}}.} \bibinfo{year}{2026}\natexlab{d}.
\newblock \bibinfo{title}{{Introducing Claude Opus 4.8}}.
\newblock \bibinfo{howpublished}{\url{https://www.anthropic.com/news/claude-opus-4-8}}.
\newblock


\bibitem[Barres et~al\mbox{.}(2025)]%
        {barres2025tau2}
\bibfield{author}{\bibinfo{person}{Victor Barres}, \bibinfo{person}{Honghua Dong}, \bibinfo{person}{Soham Ray}, \bibinfo{person}{Xujie Si}, {and} \bibinfo{person}{Karthik Narasimhan}.} \bibinfo{year}{2025}\natexlab{}.
\newblock \showarticletitle{$\tau^2$-Bench: Evaluating Conversational Agents in a Dual-Control Environment}.
\newblock \bibinfo{journal}{\emph{arXiv preprint arXiv:2506.07982}} (\bibinfo{year}{2025}).
\newblock


\bibitem[Boisvert et~al\mbox{.}(2024)]%
        {boisvert2024workarenapp}
\bibfield{author}{\bibinfo{person}{L{\'e}o Boisvert} {et~al\mbox{.}}} \bibinfo{year}{2024}\natexlab{}.
\newblock \showarticletitle{{WorkArena++}: Towards Compositional Planning and Reasoning-based Common Knowledge Work Tasks}. In \bibinfo{booktitle}{\emph{Advances in Neural Information Processing Systems (NeurIPS), Datasets and Benchmarks Track}}.
\newblock


\bibitem[Chen et~al\mbox{.}(2026)]%
        {chen2026uniclawbench}
\bibfield{author}{\bibinfo{person}{Zhekai Chen}, \bibinfo{person}{Chengqi Duan}, \bibinfo{person}{Kaiyue Sun}, \bibinfo{person}{Bohao Li}, \bibinfo{person}{Yuqing Wang}, \bibinfo{person}{Manyuan Zhang}, {and} \bibinfo{person}{Xihui Liu}.} \bibinfo{year}{2026}\natexlab{}.
\newblock \showarticletitle{{UniClawBench}: A Universal Benchmark for Proactive Agents on Real-World Tasks}.
\newblock \bibinfo{journal}{\emph{arXiv preprint arXiv:2607.08768}} (\bibinfo{year}{2026}).
\newblock


\bibitem[{DeepSeek-AI}(2026a)]%
        {deepseekai2026v4}
\bibfield{author}{\bibinfo{person}{{DeepSeek-AI}}.} \bibinfo{year}{2026}\natexlab{a}.
\newblock \showarticletitle{{DeepSeek-V4}: Towards Highly Efficient Million-Token Context Intelligence}.
\newblock \bibinfo{journal}{\emph{arXiv preprint arXiv:2606.19348}} (\bibinfo{year}{2026}).
\newblock


\bibitem[{DeepSeek-AI}(2026b)]%
        {deepseekai2026v41flash}
\bibfield{author}{\bibinfo{person}{{DeepSeek-AI}}.} \bibinfo{year}{2026}\natexlab{b}.
\newblock \showarticletitle{{DeepSeek-V4.1-Flash}: Pushing the Limits of KV Cache Compression}.
\newblock \bibinfo{journal}{\emph{arXiv preprint arXiv:2609.19969}} (\bibinfo{year}{2026}).
\newblock


\bibitem[Deng et~al\mbox{.}(2023a)]%
        {deng2023mind2web}
\bibfield{author}{\bibinfo{person}{Xiang Deng}, \bibinfo{person}{Yu Gu}, \bibinfo{person}{Boyuan Zheng}, \bibinfo{person}{Shijie Chen}, \bibinfo{person}{Samuel Stevens}, \bibinfo{person}{Boshi Wang}, \bibinfo{person}{Huan Sun}, {and} \bibinfo{person}{Yu Su}.} \bibinfo{year}{2023}\natexlab{a}.
\newblock \showarticletitle{{Mind2Web}: Towards a Generalist Agent for the Web}. In \bibinfo{booktitle}{\emph{Advances in Neural Information Processing Systems (NeurIPS), Datasets and Benchmarks Track}}.
\newblock


\bibitem[Deng et~al\mbox{.}(2023b)]%
        {deng2023proactivesurvey}
\bibfield{author}{\bibinfo{person}{Yang Deng}, \bibinfo{person}{Wenqiang Lei}, \bibinfo{person}{Wai Lam}, {and} \bibinfo{person}{Tat-Seng Chua}.} \bibinfo{year}{2023}\natexlab{b}.
\newblock \showarticletitle{A Survey on Proactive Dialogue Systems: Problems, Methods, and Prospects}. In \bibinfo{booktitle}{\emph{International Joint Conference on Artificial Intelligence (IJCAI), Survey Track}}.
\newblock


\bibitem[Drouin et~al\mbox{.}(2024)]%
        {drouin2024workarena}
\bibfield{author}{\bibinfo{person}{Alexandre Drouin} {et~al\mbox{.}}} \bibinfo{year}{2024}\natexlab{}.
\newblock \showarticletitle{{WorkArena}: How Capable are Web Agents at Solving Common Knowledge Work Tasks?}. In \bibinfo{booktitle}{\emph{International Conference on Machine Learning (ICML)}}.
\newblock


\bibitem[Froger et~al\mbox{.}(2025)]%
        {froger2025are}
\bibfield{author}{\bibinfo{person}{Romain Froger}, \bibinfo{person}{Pierre Andrews}, \bibinfo{person}{Matteo Bettini}, \bibinfo{person}{Amar Budhiraja}, \bibinfo{person}{Ricardo~Silveira Cabral}, \bibinfo{person}{Virginie Do}, \bibinfo{person}{Emilien Garreau}, \bibinfo{person}{Jean-Baptiste Gaya}, \bibinfo{person}{Hugo Lauren{\c{c}}on}, \bibinfo{person}{Maxime Lecanu}, {et~al\mbox{.}}} \bibinfo{year}{2025}\natexlab{}.
\newblock \showarticletitle{{ARE}: Scaling Up Agent Environments and Evaluations}.
\newblock \bibinfo{journal}{\emph{arXiv preprint arXiv:2509.17158}} (\bibinfo{year}{2025}).
\newblock


\bibitem[Horvitz(1999)]%
        {horvitz1999mixed}
\bibfield{author}{\bibinfo{person}{Eric Horvitz}.} \bibinfo{year}{1999}\natexlab{}.
\newblock \showarticletitle{Principles of Mixed-Initiative User Interfaces}. In \bibinfo{booktitle}{\emph{Proceedings of the SIGCHI Conference on Human Factors in Computing Systems (CHI)}}. \bibinfo{pages}{159--166}.
\newblock


\bibitem[Huang et~al\mbox{.}(2024)]%
        {huang2024crmarena}
\bibfield{author}{\bibinfo{person}{Kung-Hsiang Huang} {et~al\mbox{.}}} \bibinfo{year}{2024}\natexlab{}.
\newblock \showarticletitle{{CRMArena}: Understanding the Capacity of {LLM} Agents to Perform Professional {CRM} Tasks in Realistic Environments}.
\newblock \bibinfo{journal}{\emph{arXiv preprint arXiv:2411.02305}} (\bibinfo{year}{2024}).
\newblock


\bibitem[Huang et~al\mbox{.}(2025)]%
        {huang2025crmarenapro}
\bibfield{author}{\bibinfo{person}{Kung-Hsiang Huang}, \bibinfo{person}{Akshara Prabhakar}, \bibinfo{person}{Onkar Thorat}, \bibinfo{person}{Divyansh Agarwal}, \bibinfo{person}{Prafulla~Kumar Choubey}, \bibinfo{person}{Yixin Mao}, \bibinfo{person}{Silvio Savarese}, \bibinfo{person}{Caiming Xiong}, {and} \bibinfo{person}{Chien-Sheng Wu}.} \bibinfo{year}{2025}\natexlab{}.
\newblock \showarticletitle{{CRMArena-Pro}: Holistic Assessment of {LLM} Agents Across Diverse Business Scenarios and Interactions}.
\newblock \bibinfo{journal}{\emph{arXiv preprint arXiv:2505.18878}} (\bibinfo{year}{2025}).
\newblock


\bibitem[Jimenez et~al\mbox{.}(2024)]%
        {jimenez2024swebench}
\bibfield{author}{\bibinfo{person}{Carlos~E. Jimenez}, \bibinfo{person}{John Yang}, \bibinfo{person}{Alexander Wettig}, \bibinfo{person}{Shunyu Yao}, \bibinfo{person}{Kexin Pei}, \bibinfo{person}{Ofir Press}, {and} \bibinfo{person}{Karthik Narasimhan}.} \bibinfo{year}{2024}\natexlab{}.
\newblock \showarticletitle{{SWE}-bench: Can Language Models Resolve Real-World {GitHub} Issues?}. In \bibinfo{booktitle}{\emph{International Conference on Learning Representations (ICLR)}}.
\newblock


\bibitem[Koh et~al\mbox{.}(2024)]%
        {koh2024visualwebarena}
\bibfield{author}{\bibinfo{person}{Jing~Yu Koh}, \bibinfo{person}{Robert Lo}, \bibinfo{person}{Lawrence Jang}, \bibinfo{person}{Vikram Duvvur}, \bibinfo{person}{Ming~Chong Lim}, \bibinfo{person}{Po-Yu Huang}, \bibinfo{person}{Graham Neubig}, \bibinfo{person}{Shuyan Zhou}, \bibinfo{person}{Ruslan Salakhutdinov}, {and} \bibinfo{person}{Daniel Fried}.} \bibinfo{year}{2024}\natexlab{}.
\newblock \showarticletitle{{VisualWebArena}: Evaluating Multimodal Agents on Realistic Visual Web Tasks}. In \bibinfo{booktitle}{\emph{Annual Meeting of the Association for Computational Linguistics (ACL)}}.
\newblock


\bibitem[Li et~al\mbox{.}(2026)]%
        {li2026proevent}
\bibfield{author}{\bibinfo{person}{Guanzhen Li}, \bibinfo{person}{Liangming Pan}, {and} \bibinfo{person}{Leye Wang}.} \bibinfo{year}{2026}\natexlab{}.
\newblock \showarticletitle{{ProEvent}: An Event-centric Benchmark for Proactive Agents}.
\newblock \bibinfo{journal}{\emph{arXiv preprint arXiv:2607.17701}} (\bibinfo{year}{2026}).
\newblock


\bibitem[Liu et~al\mbox{.}(2024)]%
        {liu2024agentbench}
\bibfield{author}{\bibinfo{person}{Xiao Liu}, \bibinfo{person}{Hao Yu}, \bibinfo{person}{Hanchen Zhang}, \bibinfo{person}{Yifan Xu}, \bibinfo{person}{Xuanyu Lei}, \bibinfo{person}{Hanyu Lai}, \bibinfo{person}{Yu Gu}, \bibinfo{person}{Hangliang Ding}, \bibinfo{person}{Kaiwen Men}, \bibinfo{person}{Kejuan Yang}, {et~al\mbox{.}}} \bibinfo{year}{2024}\natexlab{}.
\newblock \showarticletitle{{AgentBench}: Evaluating {LLMs} as Agents}. In \bibinfo{booktitle}{\emph{International Conference on Learning Representations (ICLR)}}.
\newblock


\bibitem[Lu et~al\mbox{.}(2025)]%
        {lu2025proactive}
\bibfield{author}{\bibinfo{person}{Yaxi Lu}, \bibinfo{person}{Shenzhi Yang}, \bibinfo{person}{Cheng Qian}, \bibinfo{person}{Guirong Chen}, \bibinfo{person}{Qinyu Luo}, \bibinfo{person}{Yesai Wu}, \bibinfo{person}{Huadong Wang}, \bibinfo{person}{Xin Cong}, \bibinfo{person}{Zhong Zhang}, \bibinfo{person}{Yankai Lin}, \bibinfo{person}{Weiwen Liu}, \bibinfo{person}{Yasheng Wang}, \bibinfo{person}{Zhiyuan Liu}, \bibinfo{person}{Fangming Liu}, {and} \bibinfo{person}{Maosong Sun}.} \bibinfo{year}{2025}\natexlab{}.
\newblock \showarticletitle{Proactive Agent: Shifting {LLM} Agents from Reactive Responses to Active Assistance}. In \bibinfo{booktitle}{\emph{International Conference on Learning Representations (ICLR)}}.
\newblock


\bibitem[Luo et~al\mbox{.}(2025)]%
        {luo2025mcpuniverse}
\bibfield{author}{\bibinfo{person}{Ziyang Luo}, \bibinfo{person}{Zhiqi Shen}, \bibinfo{person}{Wenzhuo Yang}, \bibinfo{person}{Zirui Zhao}, \bibinfo{person}{Prathyusha Jwalapuram}, \bibinfo{person}{Amrita Saha}, \bibinfo{person}{Doyen Sahoo}, \bibinfo{person}{Silvio Savarese}, \bibinfo{person}{Caiming Xiong}, {and} \bibinfo{person}{Junnan Li}.} \bibinfo{year}{2025}\natexlab{}.
\newblock \showarticletitle{{MCP-Universe}: Benchmarking Large Language Models with Real-World Model Context Protocol Servers}.
\newblock \bibinfo{journal}{\emph{arXiv preprint arXiv:2508.14704}} (\bibinfo{year}{2025}).
\newblock


\bibitem[Merrill et~al\mbox{.}(2026)]%
        {merrill2026terminalbench}
\bibfield{author}{\bibinfo{person}{Mike~A. Merrill}, \bibinfo{person}{Alexander~G. Shaw}, \bibinfo{person}{Nicholas Carlini}, {et~al\mbox{.}}} \bibinfo{year}{2026}\natexlab{}.
\newblock \showarticletitle{{Terminal-Bench}: Benchmarking Agents on Hard, Realistic Tasks in Command Line Interfaces}.
\newblock \bibinfo{journal}{\emph{arXiv preprint arXiv:2601.11868}} (\bibinfo{year}{2026}).
\newblock


\bibitem[Mialon et~al\mbox{.}(2024)]%
        {mialon2024gaia}
\bibfield{author}{\bibinfo{person}{Gr{\'e}goire Mialon}, \bibinfo{person}{Cl{\'e}mentine Fourrier}, \bibinfo{person}{Craig Swift}, \bibinfo{person}{Thomas Wolf}, \bibinfo{person}{Yann LeCun}, {and} \bibinfo{person}{Thomas Scialom}.} \bibinfo{year}{2024}\natexlab{}.
\newblock \showarticletitle{{GAIA}: a Benchmark for General {AI} Assistants}. In \bibinfo{booktitle}{\emph{International Conference on Learning Representations (ICLR)}}.
\newblock


\bibitem[{Moonshot AI}(2026)]%
        {moonshotai2026kimik26}
\bibfield{author}{\bibinfo{person}{{Moonshot AI}}.} \bibinfo{year}{2026}\natexlab{}.
\newblock \bibinfo{title}{{Kimi K2.6}}.
\newblock \bibinfo{howpublished}{\url{https://huggingface.co/moonshotai/Kimi-K2.6}}.
\newblock


\bibitem[Nathani et~al\mbox{.}(2026)]%
        {nathani2026pare}
\bibfield{author}{\bibinfo{person}{Deepak Nathani}, \bibinfo{person}{Cheng Zhang}, \bibinfo{person}{Chang Huan}, \bibinfo{person}{Jiaming Shan}, \bibinfo{person}{Yinfei Yang}, \bibinfo{person}{Alkesh Patel}, \bibinfo{person}{Zhe Gan}, \bibinfo{person}{William~Yang Wang}, \bibinfo{person}{Michael Saxon}, {and} \bibinfo{person}{Xin~Eric Wang}.} \bibinfo{year}{2026}\natexlab{}.
\newblock \showarticletitle{Proactive Agent Research Environment: Simulating Active Users to Evaluate Proactive Assistants}.
\newblock \bibinfo{journal}{\emph{arXiv preprint arXiv:2604.00842}} (\bibinfo{year}{2026}).
\newblock


\bibitem[{OpenAI}(2026a)]%
        {openai2026gpt56}
\bibfield{author}{\bibinfo{person}{{OpenAI}}.} \bibinfo{year}{2026}\natexlab{a}.
\newblock \bibinfo{title}{{GPT-5.6}}.
\newblock \bibinfo{howpublished}{\url{https://openai.com/index/gpt-5-6/}}.
\newblock


\bibitem[{OpenAI}(2026b)]%
        {openai2026gpt54}
\bibfield{author}{\bibinfo{person}{{OpenAI}}.} \bibinfo{year}{2026}\natexlab{b}.
\newblock \bibinfo{title}{{Introducing GPT-5.4}}.
\newblock \bibinfo{howpublished}{\url{https://openai.com/index/introducing-gpt-5-4/}}.
\newblock


\bibitem[{OpenAI}(2026c)]%
        {openai2026gpt54mini}
\bibfield{author}{\bibinfo{person}{{OpenAI}}.} \bibinfo{year}{2026}\natexlab{c}.
\newblock \bibinfo{title}{{Introducing GPT-5.4 mini and nano}}.
\newblock \bibinfo{howpublished}{\url{https://openai.com/index/introducing-gpt-5-4-mini-and-nano/}}.
\newblock


\bibitem[{OpenAI}(2026d)]%
        {openai2026gpt55}
\bibfield{author}{\bibinfo{person}{{OpenAI}}.} \bibinfo{year}{2026}\natexlab{d}.
\newblock \bibinfo{title}{{Introducing GPT-5.5}}.
\newblock \bibinfo{howpublished}{\url{https://openai.com/index/introducing-gpt-5-5/}}.
\newblock


\bibitem[{OpenAI}(2026e)]%
        {openai2026gpt6solluna}
\bibfield{author}{\bibinfo{person}{{OpenAI}}.} \bibinfo{year}{2026}\natexlab{e}.
\newblock \bibinfo{title}{{Introducing GPT-6 Sol and Luna}}.
\newblock \bibinfo{howpublished}{\url{https://openai.com/index/introducing-gpt-6-sol-and-luna/}}.
\newblock


\bibitem[{OpenAI}(2026f)]%
        {openai2026gpt61sol}
\bibfield{author}{\bibinfo{person}{{OpenAI}}.} \bibinfo{year}{2026}\natexlab{f}.
\newblock \bibinfo{title}{{Introducing GPT-6.1 Sol}}.
\newblock \bibinfo{howpublished}{\url{https://openai.com/index/introducing-gpt-6-1-sol/}}.
\newblock


\bibitem[Park et~al\mbox{.}(2023)]%
        {park2023generative}
\bibfield{author}{\bibinfo{person}{Joon~Sung Park}, \bibinfo{person}{Joseph~C. O'Brien}, \bibinfo{person}{Carrie~J. Cai}, \bibinfo{person}{Meredith~Ringel Morris}, \bibinfo{person}{Percy Liang}, {and} \bibinfo{person}{Michael~S. Bernstein}.} \bibinfo{year}{2023}\natexlab{}.
\newblock \showarticletitle{Generative Agents: Interactive Simulacra of Human Behavior}. In \bibinfo{booktitle}{\emph{ACM Symposium on User Interface Software and Technology (UIST)}}.
\newblock


\bibitem[Pasternak et~al\mbox{.}(2025)]%
        {pasternak2025probe}
\bibfield{author}{\bibinfo{person}{Gil Pasternak}, \bibinfo{person}{Dheeraj Rajagopal}, \bibinfo{person}{Julia White}, \bibinfo{person}{Dhruv Atreja}, \bibinfo{person}{Matthew Thomas}, \bibinfo{person}{George Hurn-Maloney}, {and} \bibinfo{person}{Ash Lewis}.} \bibinfo{year}{2025}\natexlab{}.
\newblock \showarticletitle{Beyond Reactivity: Measuring Proactive Problem Solving in {LLM} Agents}.
\newblock \bibinfo{journal}{\emph{arXiv preprint arXiv:2510.19771}} (\bibinfo{year}{2025}).
\newblock


\bibitem[Patwardhan et~al\mbox{.}(2025)]%
        {patwardhan2025gdpval}
\bibfield{author}{\bibinfo{person}{Tejal Patwardhan}, \bibinfo{person}{Rachel Dias}, \bibinfo{person}{Elizabeth Proehl}, \bibinfo{person}{Grace Kim}, \bibinfo{person}{Michele Wang}, \bibinfo{person}{Olivia Watkins}, \bibinfo{person}{Sim{\'o}n Posada~Fishman}, \bibinfo{person}{Marwan Aljubeh}, \bibinfo{person}{Phoebe Thacker}, \bibinfo{person}{Laurance Fauconnet}, {et~al\mbox{.}}} \bibinfo{year}{2025}\natexlab{}.
\newblock \showarticletitle{{GDPval}: Evaluating {AI} Model Performance on Real-World Economically Valuable Tasks}.
\newblock \bibinfo{journal}{\emph{arXiv preprint arXiv:2510.04374}} (\bibinfo{year}{2025}).
\newblock


\bibitem[Qian et~al\mbox{.}(2024)]%
        {qian2024tellmemore}
\bibfield{author}{\bibinfo{person}{Cheng Qian}, \bibinfo{person}{Bingxiang He}, \bibinfo{person}{Zhong Zhuang}, \bibinfo{person}{Jia Deng}, \bibinfo{person}{Yujia Qin}, \bibinfo{person}{Xin Cong}, \bibinfo{person}{Zhong Zhang}, \bibinfo{person}{Jie Zhou}, \bibinfo{person}{Yankai Lin}, \bibinfo{person}{Zhiyuan Liu}, {et~al\mbox{.}}} \bibinfo{year}{2024}\natexlab{}.
\newblock \showarticletitle{Tell Me More! {T}owards Implicit User Intention Understanding of Language Model Driven Agents}. In \bibinfo{booktitle}{\emph{Annual Meeting of the Association for Computational Linguistics (ACL)}}.
\newblock


\bibitem[Qin et~al\mbox{.}(2024)]%
        {qin2024toolllm}
\bibfield{author}{\bibinfo{person}{Yujia Qin}, \bibinfo{person}{Shihao Liang}, \bibinfo{person}{Yining Ye}, \bibinfo{person}{Kunlun Zhu}, \bibinfo{person}{Lan Yan}, \bibinfo{person}{Yaxi Lu}, \bibinfo{person}{Yankai Lin}, \bibinfo{person}{Xin Cong}, \bibinfo{person}{Xiangru Tang}, \bibinfo{person}{Bill Qian}, {et~al\mbox{.}}} \bibinfo{year}{2024}\natexlab{}.
\newblock \showarticletitle{{ToolLLM}: Facilitating Large Language Models to Master 16000+ Real-world {APIs}}. In \bibinfo{booktitle}{\emph{International Conference on Learning Representations (ICLR)}}.
\newblock


\bibitem[Schick et~al\mbox{.}(2023)]%
        {schick2023toolformer}
\bibfield{author}{\bibinfo{person}{Timo Schick}, \bibinfo{person}{Jane Dwivedi-Yu}, \bibinfo{person}{Roberto Dess{\`\i}}, \bibinfo{person}{Roberta Raileanu}, \bibinfo{person}{Maria Lomeli}, \bibinfo{person}{Luke Zettlemoyer}, \bibinfo{person}{Nicola Cancedda}, {and} \bibinfo{person}{Thomas Scialom}.} \bibinfo{year}{2023}\natexlab{}.
\newblock \showarticletitle{Toolformer: Language Models Can Teach Themselves to Use Tools}. In \bibinfo{booktitle}{\emph{Advances in Neural Information Processing Systems (NeurIPS)}}.
\newblock


\bibitem[Shepard and Salimans(2026)]%
        {shepard2026automationbench}
\bibfield{author}{\bibinfo{person}{Daniel Shepard} {and} \bibinfo{person}{Robin Salimans}.} \bibinfo{year}{2026}\natexlab{}.
\newblock \showarticletitle{{AutomationBench}}.
\newblock \bibinfo{journal}{\emph{arXiv preprint arXiv:2604.18934}} (\bibinfo{year}{2026}).
\newblock


\bibitem[Styles et~al\mbox{.}(2024)]%
        {styles2024workbench}
\bibfield{author}{\bibinfo{person}{Olly Styles}, \bibinfo{person}{Sam Miller}, \bibinfo{person}{Patricio Cerda-Mardini}, \bibinfo{person}{Tanaya Guha}, \bibinfo{person}{Victor Sanchez}, {and} \bibinfo{person}{Bertie Vidgen}.} \bibinfo{year}{2024}\natexlab{}.
\newblock \showarticletitle{{WorkBench}: a Benchmark Dataset for Agents in a Realistic Workplace Setting}. In \bibinfo{booktitle}{\emph{Conference on Language Modeling (COLM)}}.
\newblock


\bibitem[Tang et~al\mbox{.}(2026)]%
        {tang2026proagentbench}
\bibfield{author}{\bibinfo{person}{Yuanbo Tang}, \bibinfo{person}{Huaze Tang}, \bibinfo{person}{Tingyu Cao}, \bibinfo{person}{Lam Nguyen}, \bibinfo{person}{Anping Zhang}, \bibinfo{person}{Xinwen Cao}, \bibinfo{person}{Chunkang Liu}, \bibinfo{person}{Wenbo Ding}, {and} \bibinfo{person}{Yang Li}.} \bibinfo{year}{2026}\natexlab{}.
\newblock \showarticletitle{{ProAgentBench}: Evaluating {LLM} Agents for Proactive Assistance with Real-World Data}.
\newblock \bibinfo{journal}{\emph{arXiv preprint arXiv:2602.04482}} (\bibinfo{year}{2026}).
\newblock


\bibitem[Trivedi et~al\mbox{.}(2024)]%
        {trivedi2024appworld}
\bibfield{author}{\bibinfo{person}{Harsh Trivedi} {et~al\mbox{.}}} \bibinfo{year}{2024}\natexlab{}.
\newblock \showarticletitle{{AppWorld}: A Controllable World of Apps and People for Benchmarking Interactive Coding Agents}. In \bibinfo{booktitle}{\emph{Annual Meeting of the Association for Computational Linguistics (ACL)}}.
\newblock


\bibitem[Wang et~al\mbox{.}(2024)]%
        {wang2024survey}
\bibfield{author}{\bibinfo{person}{Lei Wang}, \bibinfo{person}{Chen Ma}, \bibinfo{person}{Xueyang Feng}, \bibinfo{person}{Zeyu Zhang}, \bibinfo{person}{Hao Yang}, \bibinfo{person}{Jingsen Zhang}, \bibinfo{person}{Zhiyuan Chen}, \bibinfo{person}{Jiakai Tang}, \bibinfo{person}{Xu Chen}, \bibinfo{person}{Yankai Lin}, {et~al\mbox{.}}} \bibinfo{year}{2024}\natexlab{}.
\newblock \showarticletitle{A Survey on Large Language Model based Autonomous Agents}.
\newblock \bibinfo{journal}{\emph{Frontiers of Computer Science}} (\bibinfo{year}{2024}).
\newblock


\bibitem[Wang et~al\mbox{.}(2025)]%
        {wang2025odysseybench}
\bibfield{author}{\bibinfo{person}{Weixuan Wang}, \bibinfo{person}{Dongge Han}, \bibinfo{person}{Daniel Madrigal~Diaz}, \bibinfo{person}{Jin Xu}, \bibinfo{person}{Victor R{\"u}hle}, {and} \bibinfo{person}{Saravan Rajmohan}.} \bibinfo{year}{2025}\natexlab{}.
\newblock \showarticletitle{{OdysseyBench}: Evaluating {LLM} Agents on Long-Horizon Complex Office Application Workflows}.
\newblock \bibinfo{journal}{\emph{arXiv preprint arXiv:2508.09124}} (\bibinfo{year}{2025}).
\newblock


\bibitem[Wei et~al\mbox{.}(2025)]%
        {wei2025browsecomp}
\bibfield{author}{\bibinfo{person}{Jason Wei}, \bibinfo{person}{Zhiqing Sun}, \bibinfo{person}{Spencer Papay}, \bibinfo{person}{Scott McKinney}, \bibinfo{person}{Jeffrey Han}, \bibinfo{person}{Isa Fulford}, \bibinfo{person}{Hyung~Won Chung}, \bibinfo{person}{Alex~Tachard Passos}, \bibinfo{person}{William Fedus}, {and} \bibinfo{person}{Amelia Glaese}.} \bibinfo{year}{2025}\natexlab{}.
\newblock \showarticletitle{{BrowseComp}: A Simple Yet Challenging Benchmark for Browsing Agents}.
\newblock \bibinfo{journal}{\emph{arXiv preprint arXiv:2504.12516}} (\bibinfo{year}{2025}).
\newblock


\bibitem[Wu et~al\mbox{.}(2026)]%
        {wu2026asknow}
\bibfield{author}{\bibinfo{person}{Bin Wu}, \bibinfo{person}{Guanyun Zou}, \bibinfo{person}{Bingbing Wang}, \bibinfo{person}{Huan Zhao}, {and} \bibinfo{person}{Chuan Shi}.} \bibinfo{year}{2026}\natexlab{}.
\newblock \showarticletitle{Ask Now, Use Later: Benchmarking the Proactivity Gap in Long-Lived {LLM} Agents}.
\newblock \bibinfo{journal}{\emph{arXiv preprint arXiv:2605.28108}} (\bibinfo{year}{2026}).
\newblock


\bibitem[{xAI}(2026)]%
        {xai2026grok47}
\bibfield{author}{\bibinfo{person}{{xAI}}.} \bibinfo{year}{2026}\natexlab{}.
\newblock \bibinfo{title}{{Grok 4.7 Model Card}}.
\newblock \bibinfo{howpublished}{\url{https://media.x.ai/v1/website/4p7card-5eccc980.pdf}}.
\newblock


\bibitem[Xie et~al\mbox{.}(2024)]%
        {xie2024osworld}
\bibfield{author}{\bibinfo{person}{Tianbao Xie} {et~al\mbox{.}}} \bibinfo{year}{2024}\natexlab{}.
\newblock \showarticletitle{{OSWorld}: Benchmarking Multimodal Agents for Open-Ended Tasks in Real Computer Environments}. In \bibinfo{booktitle}{\emph{Advances in Neural Information Processing Systems (NeurIPS), Datasets and Benchmarks Track}}.
\newblock


\bibitem[Xu et~al\mbox{.}(2024)]%
        {xu2024theagentcompany}
\bibfield{author}{\bibinfo{person}{Frank~F. Xu}, \bibinfo{person}{Yufan Song}, \bibinfo{person}{Boxuan Li}, \bibinfo{person}{Yuxuan Tang}, \bibinfo{person}{Kritanjali Jain}, \bibinfo{person}{Mengxue Bao}, \bibinfo{person}{Zora~Z. Wang}, \bibinfo{person}{Xuhui Zhou}, \bibinfo{person}{Zhitong Guo}, \bibinfo{person}{Murong Cao}, {et~al\mbox{.}}} \bibinfo{year}{2024}\natexlab{}.
\newblock \showarticletitle{{TheAgentCompany}: Benchmarking {LLM} Agents on Consequential Real World Tasks}.
\newblock \bibinfo{journal}{\emph{arXiv preprint arXiv:2412.14161}} (\bibinfo{year}{2024}).
\newblock


\bibitem[Yang et~al\mbox{.}(2025)]%
        {yang2025contextagent}
\bibfield{author}{\bibinfo{person}{Bufang Yang}, \bibinfo{person}{Lilin Xu}, \bibinfo{person}{Liekang Zeng}, \bibinfo{person}{Kaiwei Liu}, \bibinfo{person}{Siyang Jiang}, \bibinfo{person}{Wenrui Lu}, \bibinfo{person}{Hongkai Chen}, \bibinfo{person}{Xiaofan Jiang}, \bibinfo{person}{Guoliang Xing}, {and} \bibinfo{person}{Zhenyu Yan}.} \bibinfo{year}{2025}\natexlab{}.
\newblock \showarticletitle{{ContextAgent}: Context-Aware Proactive {LLM} Agents with Open-World Sensory Perceptions}. In \bibinfo{booktitle}{\emph{Advances in Neural Information Processing Systems (NeurIPS)}}.
\newblock


\bibitem[Yang et~al\mbox{.}(2026)]%
        {yang2026fingertip}
\bibfield{author}{\bibinfo{person}{Qinglong Yang}, \bibinfo{person}{Haoming Li}, \bibinfo{person}{Haotian Zhao}, \bibinfo{person}{Xiaokai Yan}, \bibinfo{person}{Jingtao Ding}, \bibinfo{person}{Fengli Xu}, {and} \bibinfo{person}{Yong Li}.} \bibinfo{year}{2026}\natexlab{}.
\newblock \showarticletitle{{FingerTip 20K}: A Benchmark for Proactive and Personalized Mobile {LLM} Agents}. In \bibinfo{booktitle}{\emph{International Conference on Learning Representations (ICLR)}}.
\newblock


\bibitem[Yao et~al\mbox{.}(2024)]%
        {yao2024taubench}
\bibfield{author}{\bibinfo{person}{Shunyu Yao}, \bibinfo{person}{Noah Shinn}, \bibinfo{person}{Pedram Razavi}, {and} \bibinfo{person}{Karthik Narasimhan}.} \bibinfo{year}{2024}\natexlab{}.
\newblock \showarticletitle{$\tau$-bench: A Benchmark for Tool-Agent-User Interaction in Real-World Domains}.
\newblock \bibinfo{journal}{\emph{arXiv preprint arXiv:2406.12045}} (\bibinfo{year}{2024}).
\newblock


\bibitem[Yao et~al\mbox{.}(2023)]%
        {yao2023react}
\bibfield{author}{\bibinfo{person}{Shunyu Yao}, \bibinfo{person}{Jeffrey Zhao}, \bibinfo{person}{Dian Yu}, \bibinfo{person}{Nan Du}, \bibinfo{person}{Izhak Shafran}, \bibinfo{person}{Karthik Narasimhan}, {and} \bibinfo{person}{Yuan Cao}.} \bibinfo{year}{2023}\natexlab{}.
\newblock \showarticletitle{{ReAct}: Synergizing Reasoning and Acting in Language Models}. In \bibinfo{booktitle}{\emph{International Conference on Learning Representations (ICLR)}}.
\newblock


\bibitem[Yoran et~al\mbox{.}(2024)]%
        {yoran2024assistantbench}
\bibfield{author}{\bibinfo{person}{Ori Yoran}, \bibinfo{person}{Samuel~Joseph Amouyal}, \bibinfo{person}{Chaitanya Malaviya}, \bibinfo{person}{Ben Bogin}, \bibinfo{person}{Ofir Press}, {and} \bibinfo{person}{Jonathan Berant}.} \bibinfo{year}{2024}\natexlab{}.
\newblock \showarticletitle{{AssistantBench}: Can Web Agents Solve Realistic and Time-Consuming Tasks?}. In \bibinfo{booktitle}{\emph{Conference on Empirical Methods in Natural Language Processing (EMNLP)}}.
\newblock


\bibitem[Zhang et~al\mbox{.}(2026)]%
        {zhang2026pibench}
\bibfield{author}{\bibinfo{person}{Haoran Zhang}, \bibinfo{person}{Luxin Xu}, \bibinfo{person}{Zhilin Wang}, {et~al\mbox{.}}} \bibinfo{year}{2026}\natexlab{}.
\newblock \showarticletitle{$\pi$-Bench: Evaluating Proactive Personal Assistant Agents in Long-Horizon Workflows}.
\newblock \bibinfo{journal}{\emph{arXiv preprint arXiv:2605.14678}} (\bibinfo{year}{2026}).
\newblock


\bibitem[Zheng et~al\mbox{.}(2023)]%
        {zheng2023judging}
\bibfield{author}{\bibinfo{person}{Lianmin Zheng}, \bibinfo{person}{Wei-Lin Chiang}, \bibinfo{person}{Ying Sheng}, \bibinfo{person}{Siyuan Zhuang}, \bibinfo{person}{Zhanghao Wu}, \bibinfo{person}{Yonghao Zhuang}, \bibinfo{person}{Zi Lin}, \bibinfo{person}{Zhuohan Li}, \bibinfo{person}{Dacheng Li}, \bibinfo{person}{Eric~P. Xing}, {et~al\mbox{.}}} \bibinfo{year}{2023}\natexlab{}.
\newblock \showarticletitle{Judging {LLM}-as-a-Judge with {MT-Bench} and {Chatbot Arena}}. In \bibinfo{booktitle}{\emph{Advances in Neural Information Processing Systems (NeurIPS), Datasets and Benchmarks Track}}.
\newblock


\bibitem[Zhou et~al\mbox{.}(2024)]%
        {zhou2024webarena}
\bibfield{author}{\bibinfo{person}{Shuyan Zhou}, \bibinfo{person}{Frank~F. Xu}, \bibinfo{person}{Hao Zhu}, \bibinfo{person}{Xuhui Zhou}, \bibinfo{person}{Robert Lo}, \bibinfo{person}{Abishek Sridhar}, \bibinfo{person}{Xianyi Cheng}, \bibinfo{person}{Tianyue Ou}, \bibinfo{person}{Yonatan Bisk}, \bibinfo{person}{Daniel Fried}, \bibinfo{person}{Uri Alon}, {and} \bibinfo{person}{Graham Neubig}.} \bibinfo{year}{2024}\natexlab{}.
\newblock \showarticletitle{{WebArena}: A Realistic Web Environment for Building Autonomous Agents}. In \bibinfo{booktitle}{\emph{International Conference on Learning Representations (ICLR)}}.
\newblock


\end{thebibliography}

\appendix
\section{World-Building Stages}
\label{app:stages}
Table~\ref{tab:stages} lists the stages of the generator behind the four steps of Section~\ref{sec:bench-world} and Figure~\ref{fig:world}.

\subsection{Domains, Tools and Example Tasks}
Table~\ref{tab:domains} gives example tools and tasks for each domain of the simulated workspace.

\begin{table}[H]
\caption{Domains, example tools and example tasks in \benchname{}. Collaboration-tool namespace
prefixes are omitted; business-system tools are AutomationBench's \cite{shepard2026automationbench}.}
\label{tab:domains}
\begin{footnotesize}
\begin{tabular}{@{}>{\raggedright\arraybackslash}p{0.17\columnwidth}>{\raggedright\arraybackslash}p{0.42\columnwidth}>{\raggedright\arraybackslash}p{0.33\columnwidth}@{}}
\toprule
\textbf{Domain} & \textbf{Example tools} & \textbf{Example task} \\
\midrule
Collaboration & \texttt{search\_emails}, \texttt{list\_chat\_messages}, \texttt{list\_meetings},
\texttt{search\_files}, \texttt{send\_email}, \texttt{create\_document}, \texttt{share\_document} &
write the review of a completed report run and share it with its audience \\
\addlinespace[2pt]
Research & \texttt{wandb\_list\_runs}, \texttt{wandb\_run\_history}, \texttt{kubernetes\_list\_jobs},
\texttt{kubernetes\_pod\_logs}, \texttt{kubernetes\_restart\_job}, \texttt{coding\_run\_analysis} &
report which training jobs failed and why; decide which run to promote \\
\addlinespace[2pt]
Business & \texttt{zendesk\_update\_ticket}, \texttt{hubspot\_update\_contact},
\texttt{quickbooks\_send\_invoice}, \texttt{bamboohr\_update\_employee} &
update the support tickets, contacts or invoices a stated policy selects \\
\bottomrule
\end{tabular}
\end{footnotesize}
\end{table}

\subsection{Evaluation Task Families}
\label{app:taskfamilies}
Table~\ref{tab:taskfamilies} breaks down the evaluation world's tasks by family.

\begin{table}[H]
\caption{The 120 ground-truth tasks of the evaluation world by family. \emph{Evaluation} gives how the
execution is scored (Section~\ref{sec:bench-eval}).}
\label{tab:taskfamilies}
\vskip 0.15in
\begin{center}
\begin{small}
\setlength{\tabcolsep}{4pt}
\begin{tabular}{@{}>{\raggedright\arraybackslash}p{0.27\columnwidth}>{\raggedright\arraybackslash}p{0.47\columnwidth}r@{}}
\toprule
\textbf{Family} & \textbf{What the employee must do} & \textbf{Tasks} \\
\midrule
\multicolumn{2}{@{}l}{\emph{Business systems} (end state of each named record)} & 49 \\
\quad Customer support & reopen, escalate or close helpdesk tickets by a stated policy & 26 \\
\quad Sales & change CRM contacts' lifecycle stage by a stated rule & 15 \\
\quad Finance & send the invoices a stated rule selects & 5 \\
\quad Human resources & correct employee records against a roster & 3 \\
\addlinespace[2pt]
\multicolumn{2}{@{}l}{\emph{Collaboration} (document or message content)} & 54 \\
\quad Report review & write or update a report-review document & 27 \\
\quad Cross-team coordination & write or update a document on a shared work item & 15 \\
\quad Client relationship & write or update a document on a client relationship & 12 \\
\addlinespace[2pt]
\multicolumn{2}{@{}l}{\emph{Research} (delivered decision or diagnosis)} & 9 \\
\quad Promote a run & choose which training run to promote & 4 \\
\quad Restart failed jobs & restart the jobs that failed recoverably & 3 \\
\quad Triage failed jobs & report failed jobs with reasons and exit codes & 2 \\
\addlinespace[2pt]
\emph{Headcount review} & report headcount, wage bill, joiners and leavers & 8 \\
\midrule
Total & & 120 \\
\bottomrule
\end{tabular}
\end{small}
\end{center}
\vskip -0.1in
\end{table}

\begin{table*}[htbp]
\caption{The 22 world-building stages, grouped into the four steps of Figure~\ref{fig:world}. The
number after each stage is its position among the 25 invocations of a build; three stages run twice,
once more after a later stage has produced something they read. The second \texttt{build\_events},
for instance, reads the ledger's wage figures, without which headcount events are never generated.}
\label{tab:stages}
\vskip 0.15in
\begin{center}
\begin{small}
\begin{tabular*}{\textwidth}{@{\extracolsep{\fill}}>{\raggedright\arraybackslash}p{0.9in}>{\raggedright\arraybackslash}p{3.0in}>{\raggedright\arraybackslash}p{2.3in}@{}}
\toprule
\textbf{Step} & \textbf{Stages (invocation)} & \textbf{Produces} \\
\midrule
1 Company scope &
resolve\_company (1), plan\_population (2) &
a grounding profile of the company; an exact population plan \\[3pt]
2 Positions &
build\_history (3), build\_capability (4), build\_identity (5), build\_work (6), build\_context (7),
build\_access (8) &
dated employment histories; what each position does; names and accounts; projects with owners and
participants; external parties and resources; groups, roles and access grants \\[3pt]
3 Events and keys &
build\_training (9), build\_events (10, 12), build\_ledger (11), build\_graph (13),
build\_reporting (14), build\_collaboration (16), build\_cases (18), build\_query\_program (22) &
experiment runs; dated world events; payroll and numeric outcomes; a temporal graph; report runs;
collaboration episodes; anchored cases; scenarios with their oracles and keys \\[3pt]
4 Workspace &
build\_topology (15), build\_artifact\_corpus (17), build\_artifacts (19), build\_artifact\_data (20),
build\_graph (21), integrate\_artifact\_corpus (23), project\_artifact\_corpus\_graph (24),
build\_query\_program (25) &
mail, chat, calendar and file surfaces; documents with revisions and audiences; every document bound
to an exact revision and its report data; the graph and oracles rebuilt over them \\
\bottomrule
\end{tabular*}
\end{small}
\end{center}
\vskip -0.1in
\end{table*}

\section{A Successful Execution Trajectory}
\label{app:successtraj}

This section follows one research task from the request to the grade: \textsc{claude-opus-5.5} (run~1) reporting which jobs in an experiment review failed. In this and the next appendix, employee names are replaced by placeholders (person\_A, person\_B, \ldots).

{\small\raggedright
\begin{description}
\setlength{\itemsep}{2pt}\setlength{\parskip}{0pt}
\item[Request.] One-on-one chat from person\_A to person\_B, 15~Sep 2026: ``hi! this has been sitting on my side since yesterday \ldots{} About experiment review 10 --- which jobs failed, and why? Give each one's failure reason and its exit code. person\_A still hasn't had anything on it by email.''
\item[Task listed at discovery.] ``Report experiment review 10 failed jobs with failure reasons and exit codes to person\_A''; requester person\_A; evidence ``About experiment review 10 --- which jobs failed, and why? \ldots''; objective ``Identify every failed job in experiment review 10 and email person\_A each one's failure reason and exit code.''
\item[Execution (20 tool calls).]\leavevmode
\begin{itemize}
\setlength{\itemsep}{0pt}\setlength{\parskip}{0pt}
\item[1] \texttt{list\_chat\_messages} (the one-on-one chat): 50 messages.
\item[2--7] \texttt{kubernetes\_list\_jobs} and \texttt{wandb\_list\_runs} by experiment group: empty; three searches and a workspace question: the full request text.
\item[8--13] \texttt{kubernetes\_list\_jobs}, once per run and once unfiltered: five jobs, exit codes 0, 137, 0, 1, 0.
\item[14--15] \texttt{kubernetes\_get\_pod} on the two failed jobs: \texttt{OOMKilled}, exit 137; \texttt{DeadlineExceeded}, exit 1 (450\,s deadline).
\item[16] \texttt{issues\_list}: an open issue for each failed job.
\item[17--18] \texttt{search\_emails}, \texttt{get\_email}: its own earlier ``hold'' email on the same experiment group.
\item[19] \texttt{send\_email} to person\_A.
\item[20] \texttt{get\_email}: reads the sent message back.
\end{itemize}
\item[Observed change.] One email to person\_A, ``Experiment review 10: failed jobs, failure reasons and exit codes'': ``Two of the five jobs \ldots{} failed: 1.~trial-01 \ldots{} Failure reason: OOMKilled \ldots{} Exit code: 137 \ldots{} 2.~trial-03 \ldots{} Failure reason: DeadlineExceeded \ldots{} Exit code: 1 \ldots{} I haven't taken any restart action.''
\item[Grade.] The grader checks the delivery channel and recipient, and the frozen set of failed jobs with their reasons and exit codes. All matched: content recall~1, execution effect~1.
\end{description}
}

\section{Failure Trajectories}
\label{app:failtraj}

This section shows three failed runs from Table~\ref{tab:failure-examples}, in the same format, each with the step where it went wrong.

{\small\raggedright
\paragraph{Right answer, wrong channel (\textsc{gpt-5.6-sol}, run~1).}
\begin{description}
\setlength{\itemsep}{2pt}\setlength{\parskip}{0pt}
\item[Request.] The same chat message as in Appendix~\ref{app:successtraj}, ending ``person\_A still hasn't had anything on it by email.''
\item[Task listed.] ``Analyze experiment review 10 job failures''; objective ``You need to determine which experiment review 10 jobs failed and report each job's failure reason and exit code.''
\item[Execution (51 calls).] 1--6: \texttt{get\_chat}, \texttt{list\_chat\_messages} on three chats; 7--36: searches and per-run \texttt{wandb\_list\_runs}/\texttt{kubernetes\_list\_jobs}, empty under bare run ids, then five jobs under \texttt{training-run:} ids; 37--44: \texttt{kubernetes\_get\_pod}, \texttt{kubernetes\_pod\_logs} on the two failed jobs, then four searches by run id; 45: \texttt{send\_chat\_message} to the one-on-one chat; 46--51: a search and five \texttt{training\_get\_record} read-backs.
\item[Observed change.] One chat message, no email: ``For experiment review 10, 2 of the 5 jobs failed: \ldots{} OOMKilled, exit code 137. \ldots{} DeadlineExceeded, exit code 1.'' Grade~0 (no email action).
\item[Where it went wrong.] Discovery dropped ``by email'' from the objective, so the agent replied in the chat the request came from, with the same jobs, reasons and exit codes that \textsc{claude-opus-5.5} emailed.
\end{description}

\paragraph{Two requests merged into one conflict (\textsc{gpt-6-luna}, run~1).}
\begin{description}
\setlength{\itemsep}{2pt}\setlength{\parskip}{0pt}
\item[Requests.] Both from person\_C to person\_D, 9~Sep 2026. Email ``Finance system reconciliation update'': ``Send every unsent invoice whose balance is 1500 or more to its existing bill\_email, whatever its due date. \ldots'' Chat: ``Send unsent invoices with due\_date before today and a positive balance to their existing bill\_email. \ldots'' Both end ``Then email person\_C a short note confirming what you changed.''
\item[Task listed.] A single task, ``Confirm invoice criteria, send eligible invoices, and notify person\_C'', citing both messages; objective ``You need to confirm which of person\_C's conflicting invoice-selection rules controls, send only invoices covered by the confirmed rule \ldots''
\item[Execution (18 calls).] 1--13: \texttt{get\_email}, \texttt{list\_chat\_messages}, \texttt{list\_events}, searches and three workspace questions: no clarification found; 14: \texttt{quickbooks\_query} (``SELECT * FROM Invoice''): 13 invoices; 15: \texttt{send\_chat\_message} asking which rule applies; 16--18: re-reads the chat, the email and the invoices.
\item[Observed change.] One chat message: ``\ldots{} Which filter should govern? I have not sent any invoices.'' No invoice and no confirmation email were sent. Grade~0 on both requests; only one of them was matched to a listed task.
\item[Where it went wrong.] Discovery merged two separate requests into one task with a rule to confirm, and execution asked instead of carrying out each request.
\end{description}

\paragraph{An earlier reply taken as done (\textsc{gpt-5.4-mini}, run~1).}
\begin{description}
\setlength{\itemsep}{2pt}\setlength{\parskip}{0pt}
\item[Request.] Chat message from person\_A to person\_E, 6~May 2026: ``\ldots{} Compared with the 2026-03-31 one, where has the project portfolio review for 2026-04-30 got to? Put it in Model Research and Platform Engineering Operating Review notes (6 Apr 2026) \ldots'' The same day person\_A emailed ``Can you update the document for project portfolio review?''
\item[Task listed.] ``Update and send the project portfolio review document'', built on the email thread; its evidence includes person\_E's own reply on that closed thread ``Sent this over already - it went out on the day you asked.''; objective ``You update the project portfolio review document and send it back to person\_A.''
\item[Execution (5 calls).] 1: \texttt{get\_email} (the 6~May email); 2--4: \texttt{search\_emails}, \texttt{search\_files}, \texttt{search\_chats} for ``Project Portfolio Review'': nothing; 5: \texttt{get\_email} on the reply: ``Sent this over already \ldots''
\item[Observed change.] None. It reported \texttt{verified\_done}: ``\ldots{} person\_E replied on 2026-05-18 that it had already been sent on the requested day.'' Grade 0.27 (evidence partly cited; the required write was missing).
\item[Where it went wrong.] It took the employee's reply on an older, already closed thread as proof that the work was done and never opened the notes document named in the chat request.
\end{description}
}

\end{document}